\documentclass[twoside]{article}
\usepackage{graphicx}
\usepackage{caption}  
\usepackage{subcaption}
\usepackage[accepted]{aistats2023}
\usepackage{hyperref}
\usepackage{booktabs,array}
\usepackage{listings}
\usepackage{xcolor}

\usepackage[round]{natbib}

\begin{document}

% If your paper is accepted and the title of your paper is very long,
% the style will print as headings an error message. Use the following
% command to supply a shorter title of your paper so that it can be
% used as headings.
%
%\runningtitle{I use this title instead because the last one was very long}

% If your paper is accepted and the number of authors is large, the
% style will print as headings an error message. Use the following
% command to supply a shorter version of the authors names so that
% they can be used as headings (for example, use only the surnames)
%
%\runningauthor{Surname 1, Surname 2, Surname 3, ...., Surname n}

\twocolumn[

\aistatstitle{Active Predictive Coding: A Unified Neural Framework for Learning Hierarchical World Models for Perception and Planning}

\def\baselinestretch{0.98}

\aistatsauthor{ Rajesh P. N. Rao \And Dimitrios C. Gklezakos \And Vishwas Sathish}

\aistatsaddress{Paul G. Allen School of Computer Science and Engineering, University of Washington, Seattle\\
\texttt{\{rao,gklezd,vsathish\}@cs.washington.edu}} 
]

\begin{abstract}
 \vspace*{-0.1in}
  Predictive coding has emerged as a prominent model of how the brain learns through predictions, anticipating the importance accorded to predictive learning in recent AI architectures such as transformers. Here we propose a new framework for predictive coding called active predictive coding which can learn hierarchical world models and solve two radically different open problems in AI: (1) how do we learn compositional representations, e.g., part-whole hierarchies, for equivariant vision? and (2) how do we solve large-scale planning problems, which are hard for traditional reinforcement learning, by composing complex action sequences from primitive policies? Our approach exploits hypernetworks, self-supervised learning and reinforcement learning to  learn hierarchical world models that combine task-invariant state transition networks and task-dependent policy networks at multiple abstraction levels. We demonstrate the viability of our approach on a variety of vision datasets (MNIST, FashionMNIST, Omniglot) as well as on a scalable hierarchical planning problem. Our results represent, to our knowledge, the first demonstration of a unified solution to the part-whole learning problem posed by Hinton, the nested reference frames problem posed by Hawkins, and the integrated state-action hierarchy learning problem in reinforcement learning. 
  
\end{abstract}

\section{INTRODUCTION}
 \vspace*{-0.1in}
% Pred Coding intro - motivation, gene model, free energy, 
Predictive coding  \citep{Rao1999PredictiveCI,friston_pc,KELLER2018424,Jiang2021DynamicPC} has received growing attention in recent years as a model of how the brain learns models of the world through prediction and self-supervised learning. In predictive coding, feedback connections from a higher to a lower level of a cortical neural network (e.g., the visual cortex) convey predictions of lower level responses and the prediction errors are conveyed via feedforward connections to correct the higher level estimates, completing a prediction-error-correction cycle. Such a model has provided explanations for a wide variety of neural and cognitive phenomena \citep{KELLER2018424,Jiang-Rao-review}. The layered architecture of the cortex is remarkably similar across  cortical areas \citep{Mountcastle}, hinting at a common computational principle, with superficial layers (layers 2-4) receiving and processing sensory information and deeper layers (layer 5) conveying outputs to motor centers \citep{sherman}. The traditional predictive coding model focused primarily on learning visual hierarchical representations and did not acknowledge the important role of actions in learning internal world models. 

In this paper, we introduce Active Predictive Coding, a new model of predictive coding that combines state and action networks at different abstract levels to learn hierarchical internal models. The model provides a unified framework for solving three important AI problems as discussed below.

\section{RELATED WORK}
 \vspace*{-0.1in}
\noindent {\bf Part-Whole Learning Problem}. Hinton and colleagues have posed the problem of how neural networks can learn to parse visual scenes into part-whole hierarchies by dynamically allocating nodes in a parse tree. They have explored networks that use a group of neurons to represent not only the presence of an object but also parameters such as position and orientation   \citep{sabour2017dynamic,NEURIPS2019_2e0d41e0,DBLP:conf/iclr/HintonSF18,DBLP:journals/corr/abs-2102-12627}, seeking to overcome the inability of deep convolutional neural networks (CNNs) \citep{NIPS2012_c399862d} which are unable to explain the images they classify in the way humans do, in terms of objects, parts and their locations. A major open question is how neural networks can learn to create parse trees on-the-fly using learned part-whole representations.

\noindent {\bf Reference Frames Problem}. In a parallel line of research, Hawkins and colleagues  \citep{hawkins2021thousand,10.3389/fncir.2019.00022} have taken inspiration from the cortex and ``grid cells" to propose that the brain uses object-centered reference frames to represent objects, spatial environments and even abstract concepts. The question of how such reference frames can be learned and used in a nested manner for hierarchical recognition and reasoning has remained open.

\noindent {\bf Integrated State-Action Hierarchy Learning Problem}. A considerable literature exists on hierarchical reinforcement learning (see \citep{HRL-review} for a  recent survey), where the goal is to make traditional reinforcement learning (RL) algorithms more efficient through state and/or action abstraction. A particularly popular approach is to use options \citep{SUTTON1999181}, which are abstract actions which can be selected in particular states (in the option's initiation set) and whose execution executes a sequence of primitive actions as prescribed by the option's lower level policy. A major problem that has received recent attention \citep[e.g.,][]{option-critic} is learning options from interactions with the environment. The broader problem of simultaneously learning state and action abstraction hierarchies has remained relatively less explored.

 \vspace*{-0.05in}
\section{CONTRIBUTIONS OF THE PAPER}
 \vspace*{-0.1in}
The proposed active predictive coding (APC) model addresses all three problems above in a unified manner using state/action embeddings and hypernetworks \citep{DBLP:conf/iclr/HaDL17} to {\em dynamically generate and generalize} over state and action networks at multiple hierarchical levels. The APC model contributes to a number of lines of research which have not been connected before:\\ 
{\bf Perception, Predictive Coding, and Reference Frame Learning:} APC extends predictive coding and related neuroscience models of brain function \citep{Rao1999PredictiveCI,friston_pc,Jiang2021DynamicPC} to hierarchical sensory-motor inference and learning, and connects these to learning nested reference frames  \citep{hawkins2021thousand} for perception and cognition.\\
{\bf Attention Networks:}  APC extends previous hard attention approaches such as the Recurrent Attention Model (RAM) \citep{NIPS2014_09c6c378} and Attend-Infer-Repeat (AIR) \citep{NIPS2016_52947e0a} by learning structured  hierarchical strategies for sampling the visual scene.\\
{\bf Hierarchical Planning and Reinforcement Learning:} APC contributes to hierarchical planning/reinforcement learning research \citep{HRL-review,HRL-Neuro-review} by proposing a new way of simultaneously learning abstract macro-actions or options \citep{SUTTON1999181} and abstract states. \\
{\bf General Applicability in AI}: When applied to vision, the APC model learns to hierarchically represent and parse images into parts and locations. When applied to RL problems, the model can exploit hypernets to (a) define a state hierarchy, {\em not merely through state aggregation}, but by {\em abstracting transition dynamics at multiple levels}, and (b) potentially generalize learned hierarchical states and actions (options) to novel scenarios via interpolation/extrapolation in the input embedding space of the hypernetworks. 

Our approach brings us closer towards a neural solution to an important challenge in both AI and cognitive science  \citep{lake_ullman_tenenbaum_gershman_2017}: how can neural networks learn hierarchical compositional representations that allow new concepts to be  created, recognized and learned?

 \vspace*{-0.05in}
\section{ACTIVE PREDICTIVE CODING MODEL}
 \vspace*{-0.1in}
%\subsection{Basic Idea}
The APC model implements a hierarchical version of the traditional Partially Observable Markov Decision Process (POMDP) \citep{DBLP:journals/ai/KaelblingLC98,rao2010decision}, with each level employing a state network and an action network. 
%The state vector at each level is input to a hypernetwork which generates a state network for the lower level. The action vector at each level uses another hypernetwork to generate an action network (policy/option) for the lower level. The state network is a recurrent network that integrates the information from input samples at each level and implements the state transition model (mapping current state and action to next state) at a particular level of abstraction. The action network is also recurrent at each level, task-specific and produces actions for the current state at that particular level of abstraction. At every hierarchical level, the state network is trained via self-supervised learning based on predictions of inputs, while the action network is trained via reinforcement learning or planning.
 
\begin{figure}[t]
\centering
\includegraphics[width=0.8\linewidth]{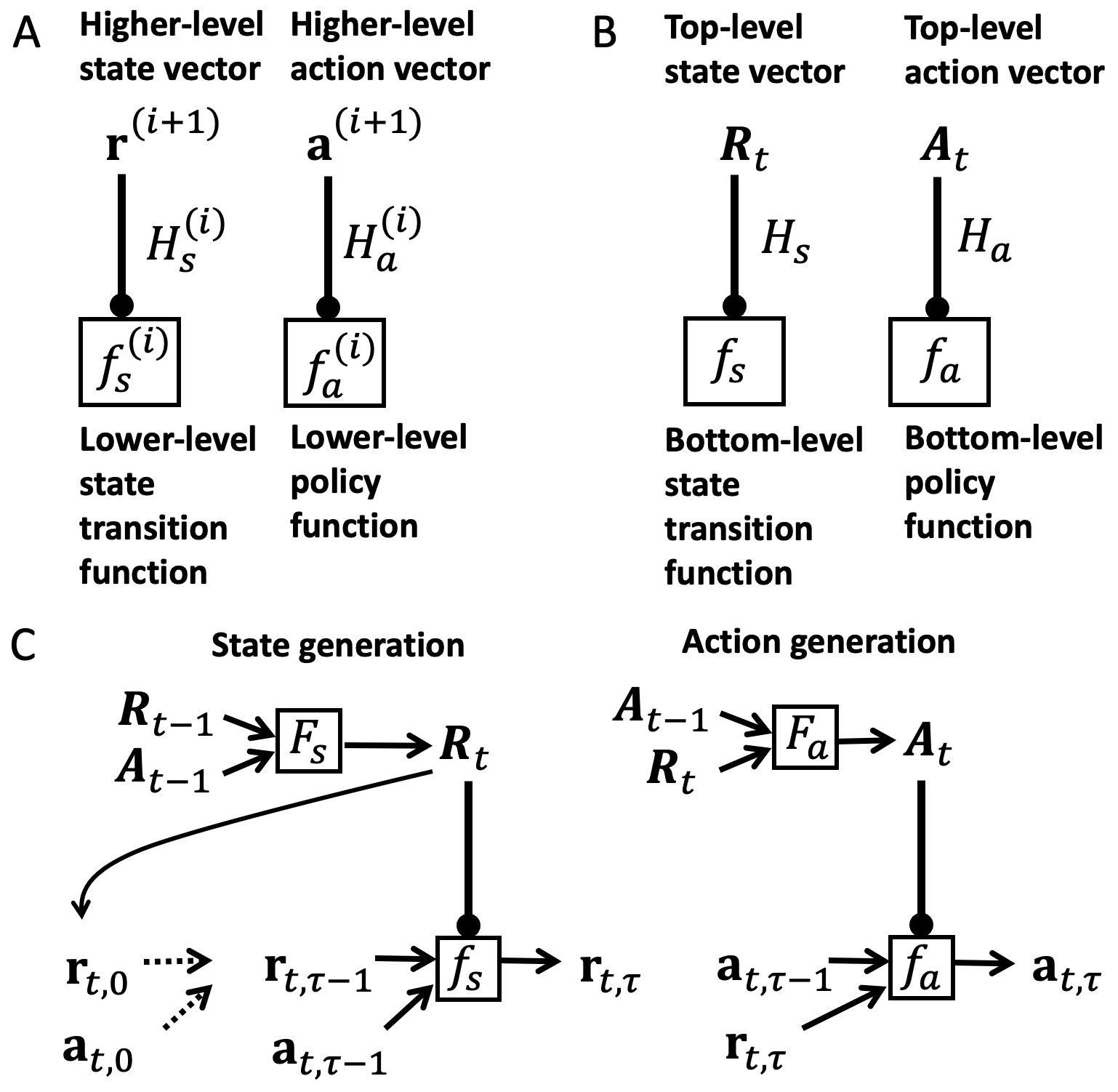}
\caption{
\textbf{Active Predictive Coding Generative Module:}
    (A) Canonical generative module for the APC model. The lower level functions are generated via hypernetworks based on the current higher level state and action embedding vectors. All functions (in boxes) are implemented as recurrent neural networks (RNNs). Arrows with circular terminations denote generation of function parameters (here, neural network weights and biases).  (B) Two-level model used in this paper. (C) Generation of states and actions in the 2-level model based on past states and actions. 
    %RNNs implementing the functions in boxes additionally receive feedback from prediction errors and lower level states/actions as described in the text.
    %\textbf{(C)} Inference and learning in a two-level HCN.  State transition and policy/action functions are implemented using recurrent neural networks (RNNs). Hypernets generate lower level state and action recurrent neural networks based on current high level state and action vectors. These recurrent networks utilize prediction errors to correct their state/action estimates as they generate parts/subparts within the intrinsic reference frame defined by the higher level state vector. Prediction errors are also used to train the state transition networks. Reinforcement learning based on the task loss (here, classification loss) and the REINFORCE algorithm is used to train the action networks.
    }
    \vspace*{-0.15in}
\label{Fig:APCN-intro}
\end{figure}

Figure~\ref{Fig:APCN-intro}A shows the canonical  generative module for the APC model. The module consists of (1) a {\em higher level state embedding vector} ${\bf r}^{(i+1)}$ at level $i+1$, which uses a function $H^{i}_s$ (implemented as a {\em hypernetwork} \citep{DBLP:conf/iclr/HaDL17})\footnote{See Supplementary Materials for an alternate neural implementation of the APC model using higher-level embedding-based inputs to lower-level RNNs.}  to generate a lower level state transition {\em function} $f^{i}_s$ (implemented as an RNN), and (2) a {\em higher level action embedding vector} ${\bf a}^{(i+1)}$, which uses a function (hypernetwork) $H^{i}_a$ to generate a lower level {\em option/policy function} $f^{i}_a$ (implemented as an RNN).  
 The state and action networks at the lower level are generated independently (by the higher level state/action embedding vectors) but exchange information horizontally within each level as shown in Figure~\ref{Fig:APCN-intro}C: the state network generates the next state prediction based on the current state and action while the action network generates the current action based on the current state and previous action.  In our current implementation, the lower level RNNs execute for a fixed number of time steps before returning control back to the higher level.\footnote{Future implementations will explore the use of termination functions \citep{SUTTON1999181,NIPS2016_52947e0a} to allow a variable number of time steps at each level and for each input.}  
 For the present paper, we focus on a two-level model (with a top level and bottom level) as shown in Figures~\ref{Fig:APCN-intro}B-C.

 %The higher level state then transitions to the next state using that level's transition function $F_s$ and the action specified by the higher level policy $F_a$, and the process continues.

%The state transition function $f^{i}_s$ takes as input the current state $r^{i}_t$ at that level at time step $t$ and the current action $a^{i}_t$ to generate the next state $r^{i}_{t+1}$ (Figure~\ref{Fig:HCN}B, "State generation"). The initial state $r^{i}_0$ is generated by the higher level state $\bf{r}^{(i+1)}$. The action network takes as input the current state $r^{i}_t$ (and optionally, the previous action $a^{i}_{t-1}$) to generate the current action $a^{i}_{t}$ (Figure~\ref{Fig:HCN}B, "Action generation"). 

 \vspace*{-0.05in}
\subsection{Inference in the Active Predictive Coding Model}
 \vspace*{-0.1in}
Inference involves estimating the state and action vectors at multiple levels based on the sequence of inputs produced by interacting with the environment in the context of a particular task or goal.  
%Without loss of generality, we consider here a two-level version of the model. The two levels operate at different time scales. 
The top level runs for $T_2$ steps (referred to as ``macro-steps''). For each macro-step, the bottom level runs for $T_1$ ``micro-steps''. As shown in Figure~\ref{Fig:APCN-intro}C, $F_s,F_a$ are the top level state and action networks respectively, and  $R_{t},A_{t}$ are the recurrent activity vectors of these networks (i.e., the top level state and action embedding vectors) at macro-step $t$. We use the notation $f(;\theta)$ to denote a network parameterized by $\theta = \{W_l,b_l\}_{l=1}^{L}$, the weight matrices and biases for all the layers. The bottom level state and action RNNs are denoted by $f_s(;\theta_{s})$ and $f_a(;\theta_{a})$, while their activity vectors are denoted by $r_{t,\tau}$ and $a_{t,\tau}$ respectively ($t$ ranges over macro-steps,  $\tau$ over micro-steps). 

\begin{figure}[t]
\centering
\includegraphics[width=0.75\linewidth]{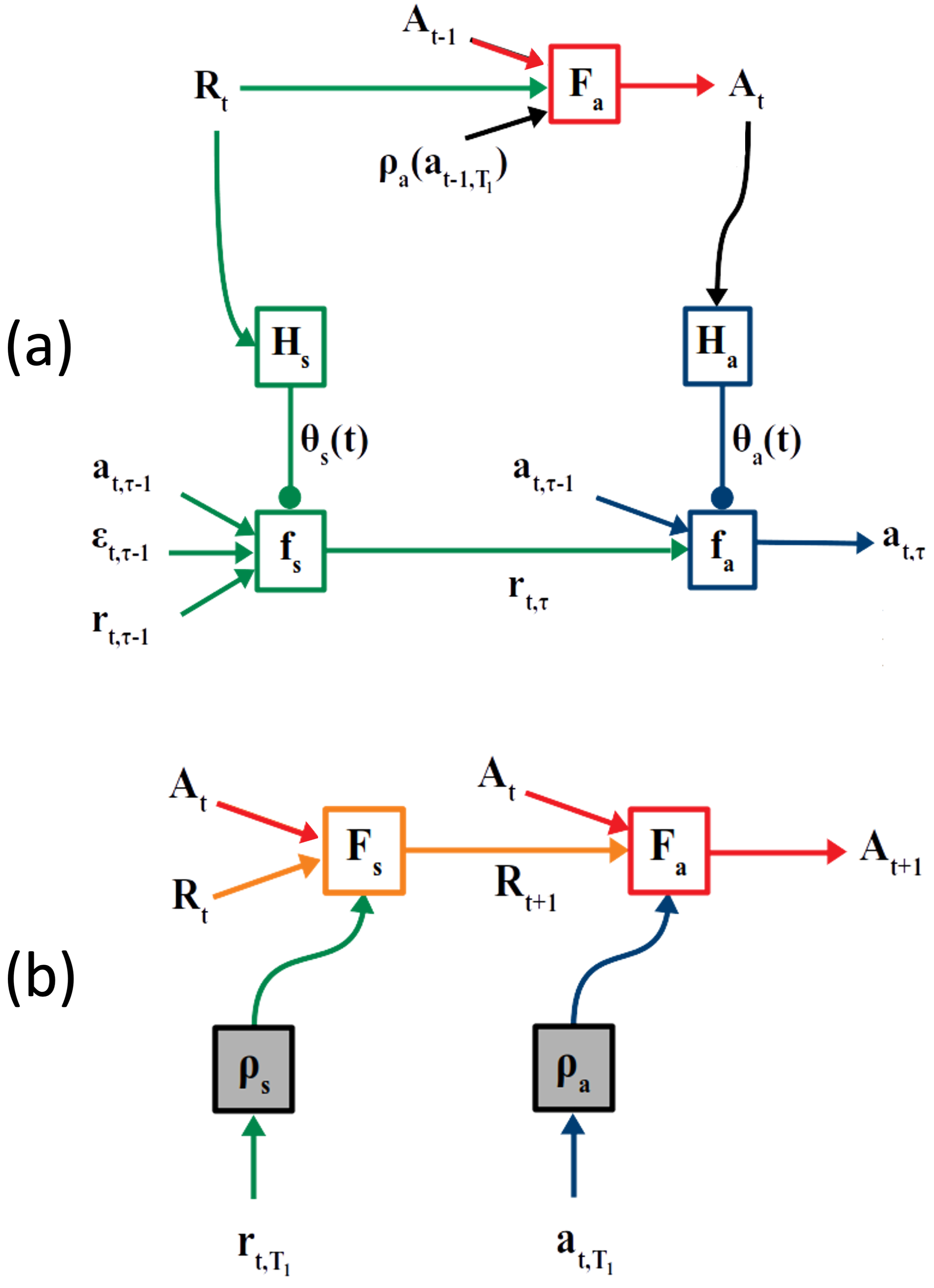}

    \caption{
        \textbf{Inference in the Active Predictive Coding Model:} (a) Dynamic generation of bottom-level state RNN $f_s$ and action RNN $f_a$ (``sub-programs'') from top-level state vector $R_t$ and action vector $A_t$. This diagram elaborates the one in Figure~\ref{Fig:APCN-intro}C.  (b) Update of top-level state $R_t$ and action $A_t$ based on feedback (via networks $\rho_s$ and $\rho_a$) upon bottom-level sub-program termination.
    }
        %\label{fig:main-hyper}
        %\label{fig:main-feedback}
\label{fig:main}
\end{figure}

 \vspace*{-0.05in}
\subsubsection{Higher-Level Inference and Reference Frame Generation}
 \vspace*{-0.1in}
At each macro-step $t$, the top level state RNN $F_s$ produces a new state embedding vector $R_t$ based on the previous state and action embedding vectors. This higher level state $R_t$ defines a new ``reference frame'' for the lower level to operate over as follows: $R_t$ is fed as input to the state hypernetwork $H_s$ to generate the lower level parameters $\theta_{s}(t) = H_s(R_t)$ specifying a dynamically generated bottom-level state RNN characterizing the state transition dynamics locally (e.g., local parts and their transformations in vision (see Section~\ref{vision}), navigation dynamics in a local region of a building (see Section~\ref{maze})). Figure \ref{fig:main} (a) illustrates this top-down generation process.

The current state $R_t$ is also input to the action/policy RNN $F_a$ which outputs an action embedding vector $A_t$ (a macro-action/option/sub-goal) appropriate for the current task/goal given current state $R_t$ (Figure~\ref{fig:main} (a)). This embedding vector $A_t$ is used as input to a non-linear function, implemented by the hypernetwork $H_a$, to dynamically generate the parameters $\theta_{a}(t) = H_a(A_t)$ of the lower-level action RNN, which implements a  policy to generate primitive actions suitable for achieving the sub-goal associated with $A_t$. Since reinforcement learning or planning require exploration, the output of an action network at any level can be regarded as the mean value of a Gaussian with fixed variance \citep{NIPS2014_09c6c378} or as a categorical representation \citep{Hafner-dreamerv2} to sample an action. 

 \vspace*{-0.05in}
\subsubsection{Lower-Level Inference and Interaction with the Higher Level}
 \vspace*{-0.1in}
At the beginning of each micro-step, the higher-level state $R_t$ is used to initialize the bottom-level state vector via a small feedforward network $\text{Init}_s$ to produce $r_{t,0} = \text{Init}_s(R_t)$. Each micro-step proceeds in a manner similar to a macro-step. The bottom-level action RNN, which was generated by higher-level state $R_t$, produces the current action $a_{t,\tau}$ based on the current lower level state and previous action (Figure \ref{fig:main} (a) lower right). This action (e.g., eye movement or body movement) results in a new input being generated by the environment for the bottom (and possibly higher) state network. 

To predict this new input,  the lower-level state vector $r_{t,\tau}$ is fed to a generic decoder network $D$ to generate the prediction $\hat{I}_{t,\tau}$. This predicted input is compared to the actual input to generate a prediction error $\epsilon_{t,\tau} = I_{t,\tau} - \hat{I}_{t,\tau}$. Following the predictive coding model \citep{Rao1999PredictiveCI}, the prediction error is used to update the state vector via the state network: $r_{t,\tau+1} = f_s(r_{t,\tau},\epsilon_{t,\tau},a_{t,\tau};\theta_{(s)}(t))$ (Figure \ref{fig:main} (a)  lower left). 
%The lower level follows the same Gaussian noise-based exploration strategy for sampling locations as the top-level.

At the end of each macro-step (after $T_1$ bottom-level micro-steps have finished executing), the top level state RNN activity vector is updated using the final bottom-level state vector: $R_{t+1} = F_s(R_t,A_t,\rho_s(r_{t,T_1}))$ where $\rho_s()$ is a single-layer state ``feedback'' network (see Figure~\ref{fig:main} (b) left side). The top-level action RNN then produces the action vector $A_{t+1}$ based on state $R_{t+1}$ and lower level feedback $\rho_a(a_{t,T_1})$ (Figure~\ref{fig:main} (b) right side), and the process continues. The above steps correspond to a sub-program in the state/action hierarchy terminating and returning its result to be integrated by its parent. Note that more levels can be added by having $F_s,F_a$ be dynamically generated by another parent level, and so on.

 \vspace*{-0.05in}
\subsection{Training the Active Predictive Coding Model}
 \vspace*{-0.1in}
Since the state networks are task-agnostic and geared towards capturing the dynamics of the world, they are trained using self-supervised learning by minimizing prediction errors (here, via backpropagation). The action networks are trained to minimize total expected task loss: this can be done using either reinforcement learning or planning with the help of the state networks. In Section~\ref{vision} (learning equivariant vision), we illustrate the use of the REINFORCE algorithm \citep{REINFORCE-ref} for training the action networks while in Section~\ref{maze} (learning navigation), we illustrate the use of planning but note that the APC framework is flexible and allows either approach for estimating actions. 
%The goal of the state prediction network is to predict the next state and is task-agnostic. The goal of the action network is to choose effective actions given past states and actions, so that the task loss is minimized (see Supplementary Materials).

\section{APPLICATION TO EQUIVARIANT VISION AND PART-WHOLE LEARNING}
\label{vision}
 \vspace*{-0.1in}
A long standing problem in computer vision and cognitive science is: how can neural networks learn intrinsic references frames for objects and parse visual scenes into part-whole hierarchies by dynamically allocating nodes in a parse tree \citep{DBLP:journals/corr/abs-2102-12627}? Human vision provides an important clue. Unlike CNNs which process an entire scene, human vision is an active sensory-motor process,  sampling the scene via eye movements to move the high-resolution fovea to appropriate locations and accumulating evidence for or against competing visual hypotheses \citep{human-eye-movements-review}. As described below, the APC model is well-suited to emulating the sensory-motor nature of human vision, given its integrated state and action networks.

 \vspace*{-0.05in}
\subsection{APC Solution to the Equivariant Vision and Part-Whole Learning Problem}
 \vspace*{-0.1in}
 For equivariant vision and part-whole learning, the actions in the APC model emulate eye movements (or ``attention'') and correspond to moving a ``glimpse sensor'' \citep{NIPS2014_09c6c378} which extracts high-resolution information about a small part of a larger input image. 
Given an input image $I$ (of size $N\times N$ pixels), this sensor, $G$, takes in a location $l$ and a fixed scale fraction $m$, and extracts a square glimpse/patch $g = G(I,l,m)$ centered at $l$ and of size $(mN)\times(mN)$. The location $l$ to fixate next is selected by the bottom level action network within the reference frame specified by the higher level, while the bottom state network generates a prediction of the glimpse, the prediction error being used for updating the state and for learning.

%Parsing an image using the APC model corresponds to the following: the current action vector at a given level effectively specifies which sub-tree of the image parse tree to explore next, while the current state vector represents all the integrated image information provided by the lower levels. The exploration of a sub-tree proceeds by dynamically generating state and action RNNs (``programs'') for the level below via hypernetworks. These RNNs act as ``programs'' to predict and parse the lower level parts of the input image via sequences of locations/transformations computed by the action RNN, before returning control back to the higher level. The higher level state then transitions to the next state (the next object/part) using transition function $F_s$ and the next action specified by policy $F_a$.

In more detail, at each macro-step $t$, the top level action vector $A_t$ generates two values: (a) a location $L_t$ and (b) a macro-action (or option) $z_t$. The location $L_t$ is used to restrict the bottom level to a sub-region $I^{(1)}_t = G(I,L_t,M)$ of scale $M$ centered around $L_{t}$, corresponding to a new frame of reference selected by the top level within the input image. The option $z_t$, which operates over this frame of reference, is used as an embedding vector input to the hypernetwork $H_a$ to generate the parameters of the bottom level action RNN. For exploration during reinforcement learning, we treat the output of the location network as a mean value $\bar{L}_t$ and add Gaussian noise with fixed variance to sample an actual location: $L_t = \bar{L}_t + \epsilon$, where $\epsilon \sim \mathcal{N}(0,\sigma^2)$. We do the same for the option $z_t$.

Based on the current bottom-level state and action, the bottom-level action RNN generates a new action $a_{t,\tau}$. A location $l_{t,\tau}$ is chosen as a function of $a_{t,\tau}$, resulting in a new glimpse image $g_{t,\tau} = G(I^{(1)}_t,l_{t,\tau},m)$ of scale $m$ centered around $l_{t,\tau}$ and yielding a nested reference frame within the larger reference frame of $I^{(1)}_t$ specified by the higher level. %The frames of reference and the corresponding image sub-regions across the two levels are depicted in Figure \ref{fig:frames}.
The bottom level follows the same Gaussian noise-based exploration strategy for sampling locations as the top level.
The bottom-level state vector $r_{t,\tau}$, along with locations $L_t$ and $l_{t,\tau}$, are fed to a generic decoder network $D$ to generate the predicted glimpse $\hat{g}_{t,\tau}$. Following the predictive coding model \citep{Rao1999PredictiveCI}, the resulting prediction error $\epsilon_{t,\tau} = g_{t,\tau} - \hat{g}_{t,\tau}$ is used to update the state vector: $r_{t,\tau+1} = f_s(r_{t,\tau},\epsilon_{t,\tau},l_t;\theta_{(s)}(t))$. For the results below, the state networks at both levels were trained to minimize image  prediction errors while the action networks were trained using reinforcement learning (the REINFORCE algorithm \citep{REINFORCE-ref}).

 \vspace*{-0.1in}
\subsection{Results}
\label{sec:datasets}
 \vspace*{-0.1in}
We first tested the APC model on the task of  sequential part/location prediction and image reconstruction of objects in the following datasets: (a) MNIST: Original MNIST dataset of 10 classes of handwritten digits.
    %\item Rot-MNIST: MNIST digits randomly rotated at $10$ distinct angles equally spaced between $[-\frac{\pi}{2},\frac{\pi}{2}]$.
(b) Fashion-MNIST: Instead of digits, the dataset consists of $10$ classes of clothing items.
(c) Omniglot: $1623$ hand-written characters from $50$ alphabets, with $20$ samples per character. For our APC models, we used $3$ macro- and $3$ micro-steps (except $4$ macro-steps for Omniglot). A single dense layer, together with an initial random glimpse was used to initialize the state and action vectors of the top level (see Supplementary Materials for details).

\begin{figure}
\centering
\includegraphics[width=1.01\linewidth]{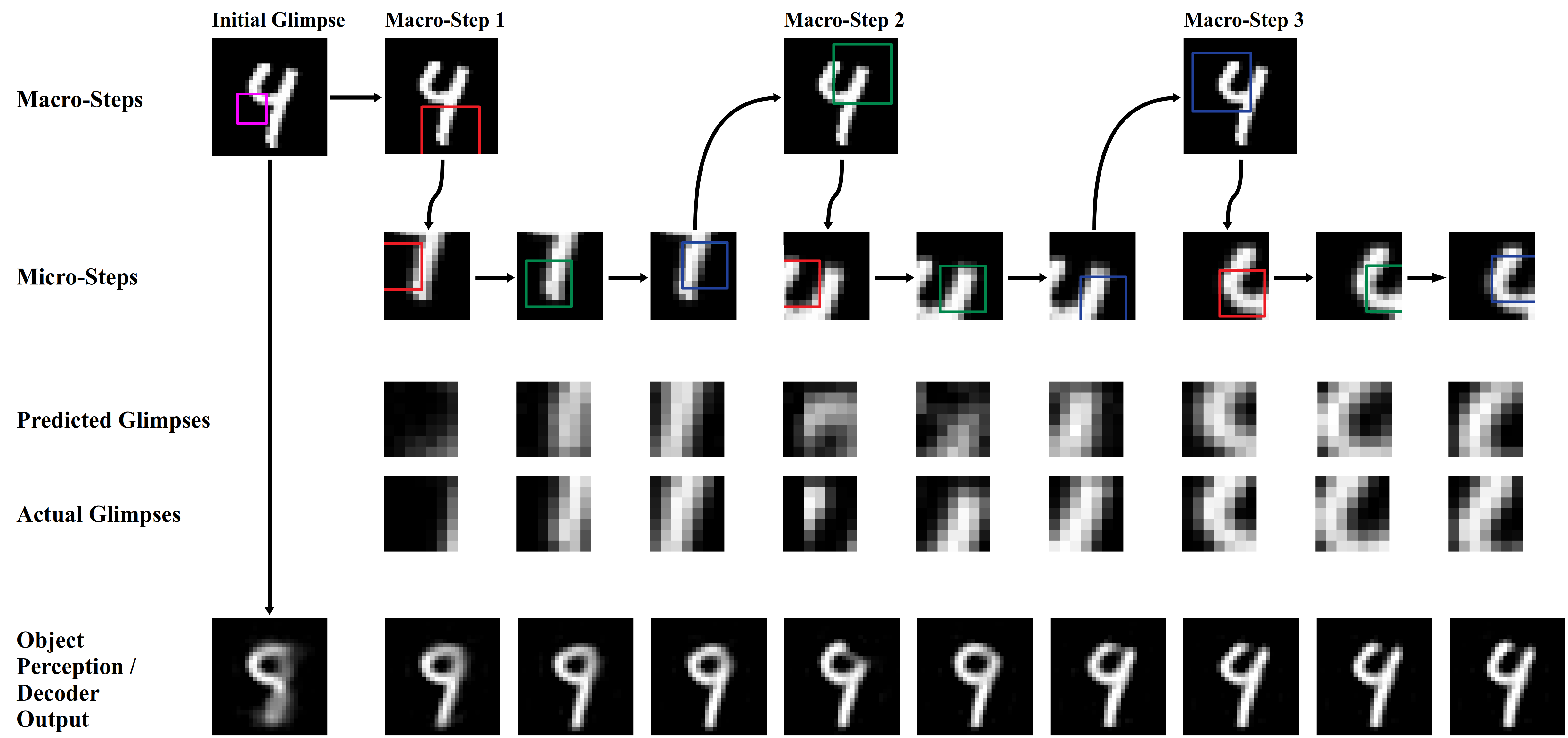}
\caption{
\textbf{Learned Two-Level Parsing Strategy and an Illustration of Perceptual Stability in the presence of Eye Movements:} 1st row: Initialization glimpse (purple box) and sampled top-level reference frames (red, green, blue boxes), 2nd row: Sampled bottom level parts within each top-level frame, 3rd \& 4th rows: Predicted versus actual parts/glimpses, and 5th row: ``Perception'' of the model (object reconstructed from current network state) over time. Note the perceptual stability in the model despite jumps in the sampled glimpses (see row of Actual Glimpses) as the object hypothesis is gradually refined based on accumulated evidence.  
}
\label{fig:parse-example}
\end{figure}

{\bf Learning to Parse Images into Parts and Perceptual Stability}. 
Figure \ref{fig:parse-example} shows an example of a learned parsing strategy by a two-level APC model for an MNIST digit. The top-level learned to cover the input image sufficiently while the bottom level learned to parse sub-parts inside the reference frame computed by the higher level. 

Figure \ref{fig:parse-example} also provides an explanation for why human perception can appear stable despite dramatic changes in our retinal images as our eyes move to sample a scene: the last row of the figure shows how the model maintains a visual hypothesis that is gradually refined and which does not exhibit the kind of rapid changes seen in the sampled images (``Actual Glimpses" in Figure~\ref{fig:parse-example}). 

\begin{figure}
\centering
\begin{subfigure}{\linewidth}
 \centering
  \includegraphics[width=0.75\linewidth]
  {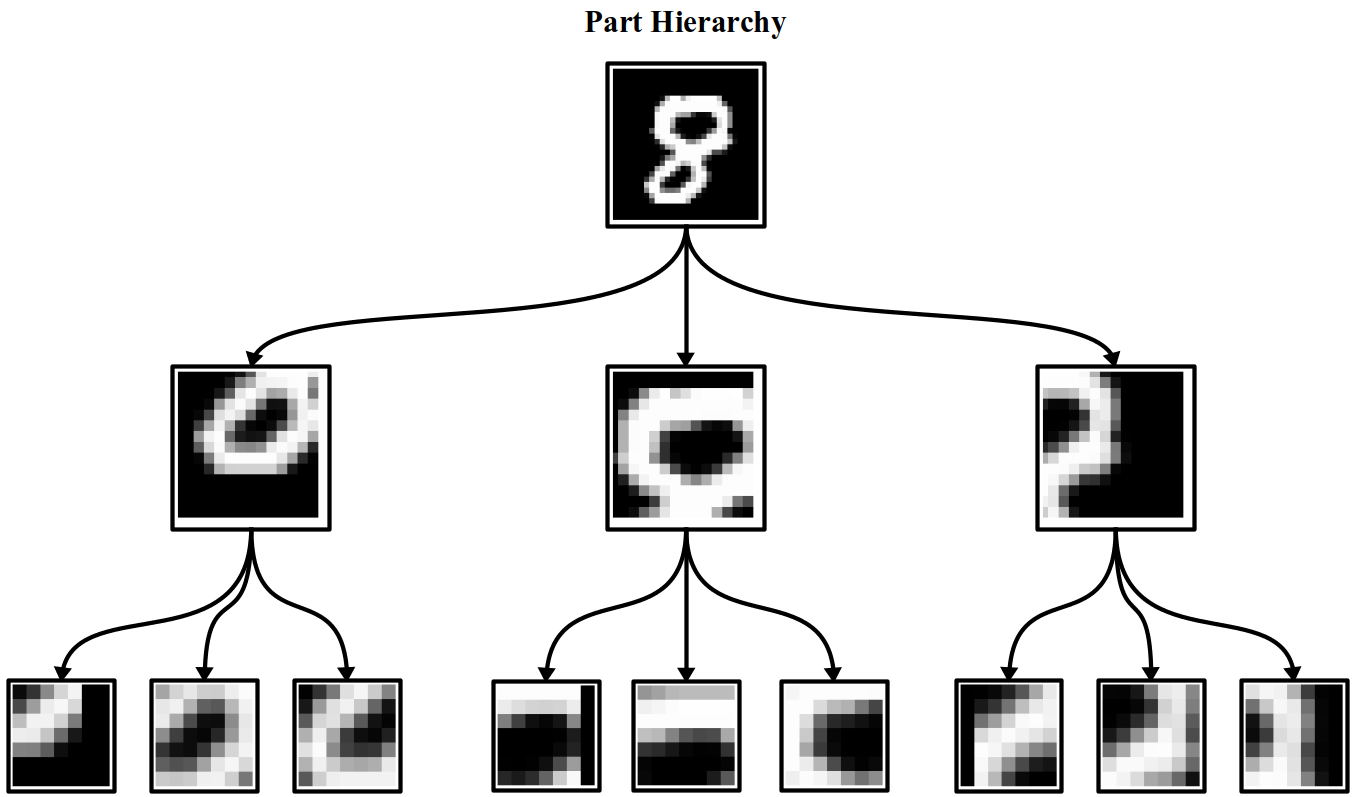}
    \caption{}
    \label{fig:parts-hierarchy}
\end{subfigure}
~
\begin{subfigure}{\linewidth}
 \centering
  \includegraphics[width=0.75\linewidth]
  {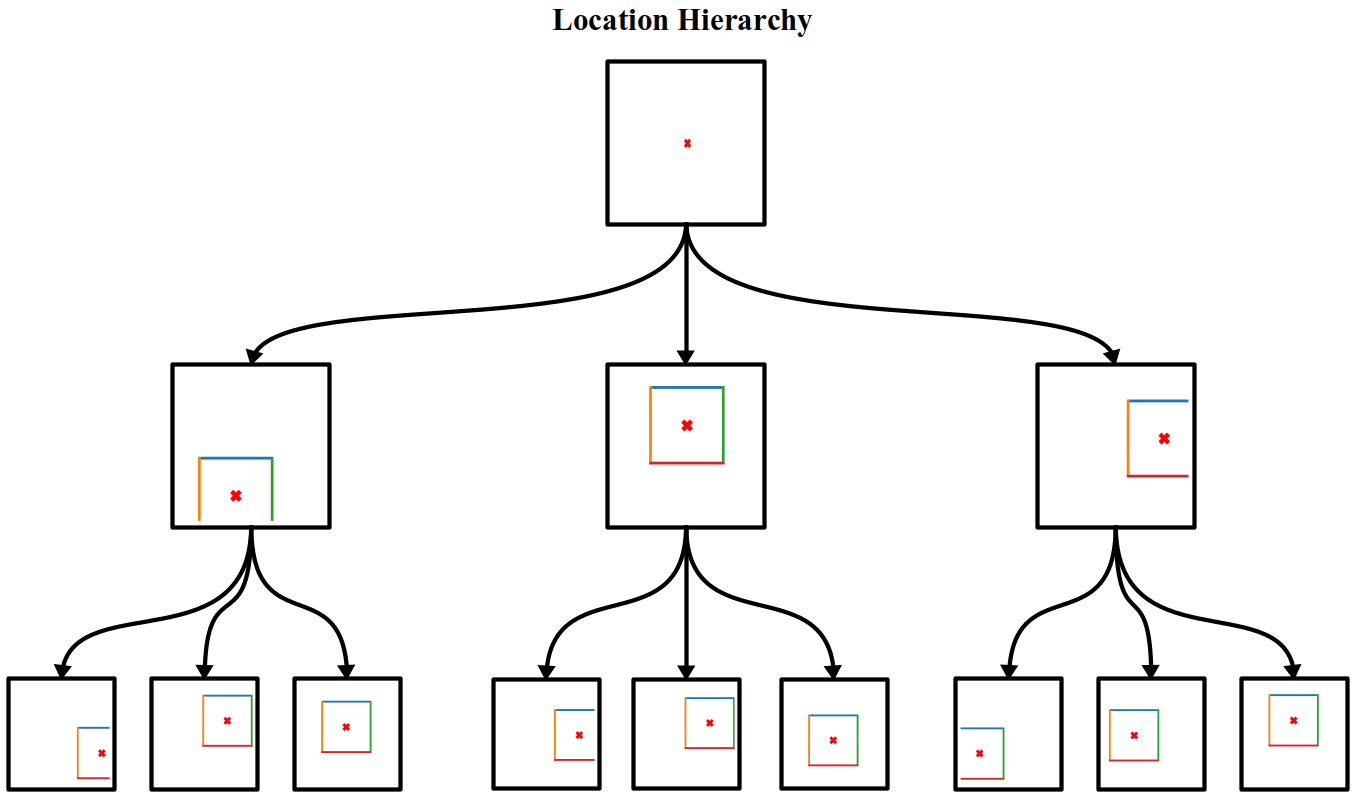}
    \caption{}
    \label{fig:locs-hierarchy}
\end{subfigure}
\caption{
\textbf{Example Parse Tree with Inferred Locations of Parts:} Hierarchy of (a) sampled parts and (b) sampled locations, inducing a hierarchy of reference frames.
}
\label{fig:parse-tree-example}
\end{figure}

\begin{figure}
\centering
\begin{subfigure}{0.75\linewidth}
\centering
\includegraphics[width=\linewidth]{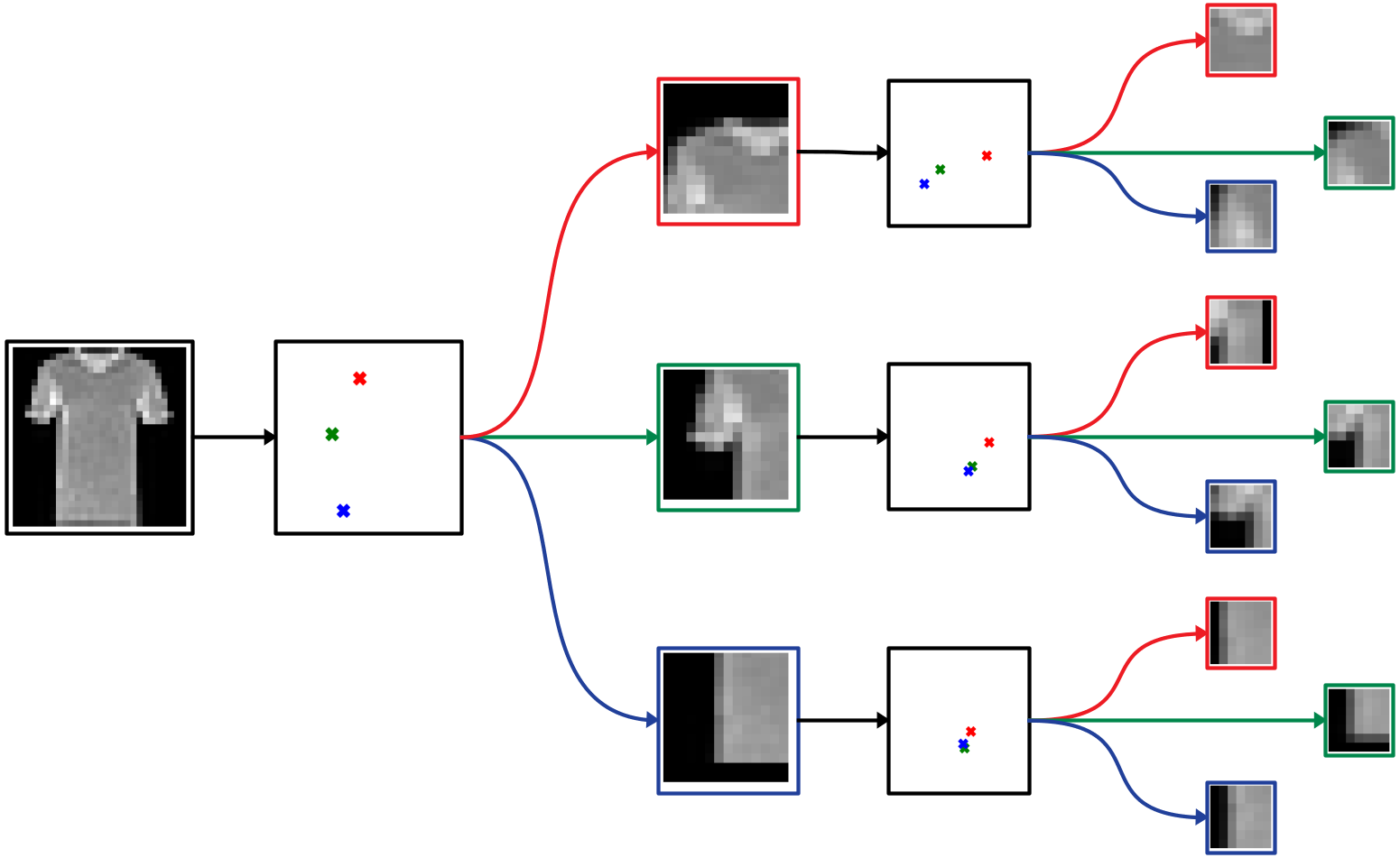}
    \caption{}
    \label{fig:class-tshirt}
\end{subfigure}\\
\begin{subfigure}{0.75\linewidth}
\centering
\includegraphics[width=\linewidth]{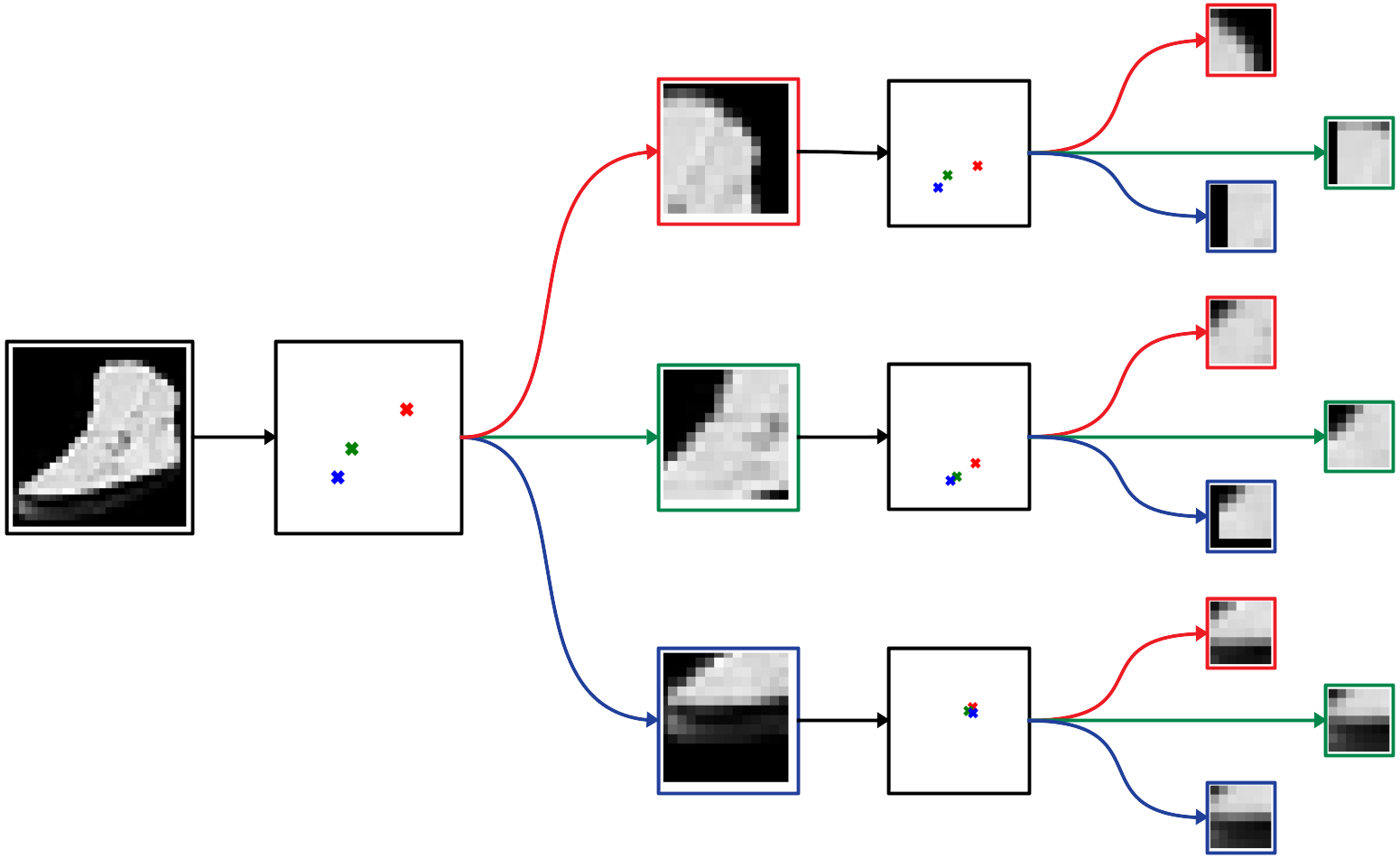}
    \caption{}
    \label{fig:class-sneaker}
\end{subfigure}\\
\centering
\caption{
\textbf{Class-Based Hierarchical Representation of Object Parts and Locations:} Average sampled locations per class, together with sampled parts for one specific example for (a) the T-shirt and (b) the sneaker classes. The order of sampled locations within each frame of reference is 1st: red, 2nd: green and 3rd: blue.
}
\label{fig:class-based-parts}
\end{figure}

\begin{figure}[t]
% \begin{subfigure}{\textwidth}
% \includegraphics[width=\linewidth]{figs/mnist_clusters.png}
%     \caption{}
%     \label{fig:parts-hierarchy}
% \end{subfigure}
%\begin{subfigure}{\textwidth}
\includegraphics[width=0.9\linewidth]{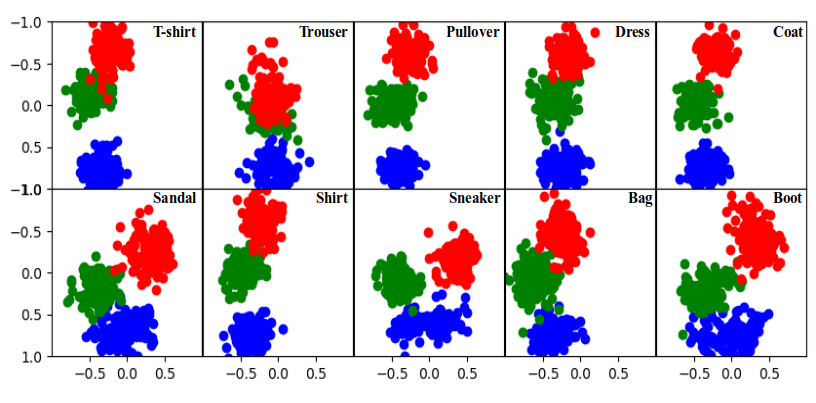}
    %\caption{}
    %\label{fig:locs-hierarchy}
%\end{subfigure}
\centering
\vspace*{-0.05in}
\caption{
\textbf{Top-level Part Locations for Fashion-MNIST Examples by Class:} Red: first, Green: second and Blue: third location. Note the differences in top-level action strategies between vertically symmetric items (shirts, trousers, bags) and footwear (sandals, sneakers, boots).
}
\label{fig:top-parts-FMNIST}
\end{figure}

Figure~\ref{fig:parse-tree-example} shows a learned part-whole hierarchy for an MNIST input, in the form of a parse tree of parts and sub-parts (strokes and mini-strokes) with their locations. The model learns different parsing strategies and different part-whole hierarchies for different classes of objects, as seen in Figure~\ref{fig:class-based-parts} for two different clothing items from the Fashion-MNIST dataset (T-shirt versus sneaker). Top-level part locations across all classes are shown in Figure~\ref{fig:top-parts-FMNIST}.

\begin{figure} [t]
\centering
    \begin{subfigure}{0.4\linewidth}
    \centering
\includegraphics[width=\linewidth]
{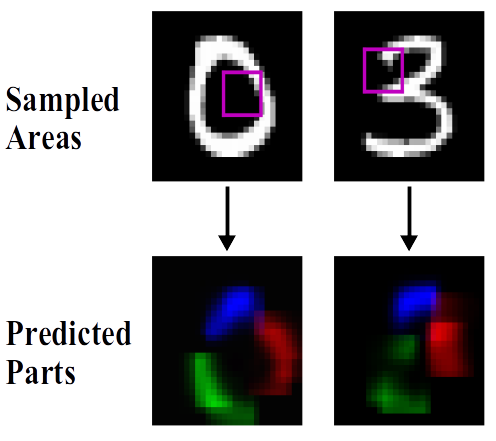}
\caption{}
        \label{fig:MNIST-parts}
    \end{subfigure} 
    \hfill
    \begin{subfigure}{0.4\linewidth}
    \centering
        \includegraphics[width=\linewidth]
        {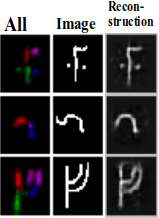}
        \caption{}
        \label{fig:omni-transfer}
    \end{subfigure}
\vspace*{-0.05in}
\caption{
\textbf{Prediction of Parts, Pattern Completion and Transfer Learning:} (a) Given only an initialization glimpse (purple box) for an input image (here, a 0 and a 3 from MNIST), an APC model trained on MNIST predicts its best guess of the parts of the object and their locations (colored segments in row below). (b) APC model trained on Omniglot can transfer its learned knowledge to predict parts of previously unseen character classes. First column: all the predicted parts. Middle column: input from a novel class. Last column: APC model reconstruction.
}
%\textbf{Prediction and Pattern Completion by APCNs:} Given an incomplete input, APCN predicts rest of the object by hallucinating object locations with large prediction errors and replacing those zeros (similar to robust predictive coding). MNIST, FMNIST, Omniglot examples.

\label{fig:part-prediction-and-transfer}
\end{figure}

{\bf Prediction of Parts and Pattern Completion}. To investigate the predictive and generative ability of the model, we had the model ``hallucinate'' different parts of an object by setting the prediction error input to the lower level network to zero. This disconnects the model from the input, forcing it to predict the next sequence of parts and ``complete'' the object. Figure~\ref{fig:MNIST-parts} shows that the model has learned to generate plausible predictions of parts given an initial glimpse.

{\bf Transfer Learning}. We tested transfer learning for reconstruction of unseen character classes for the Omniglot dataset. We trained a two-level APC model to reconstruct examples from $85\%$ of classes from each Omniglot alphabets. The rest of the classes were used to test transfer: the trained model had to generate new  programs (via the state and action hypernets) to predict parts for new character classes for each alphabet. The model successfully performed this transfer task (Table \ref{table:rec-mse} and Figure~\ref{fig:omni-transfer}).  

{\bf Ablation Studies}. To test the utility of having two levels of abstraction, we compared the image reconstruction performance of the two-level APC model (APC-2) to a one-level model (APC-1). To test the importance of intelligent sampling of the image, we used a Randomized Baseline model RB, which samples glimpses (same size glimpses as APC-1 and APC-2) from $T$ i.i.d. locations ($T$ same as for APC-1 and APC-2, i.e., 9 for MNIST/FMNIST, 12 for Omniglot), extracts an average feature vector and feeds this to a feedforward network to reconstruct the image. As shown in Table \ref{table:rec-mse}, APC-2 outperforms or matches APC-1 and RB. APC-2 also outperforms RB and APC-1 on the Ommiglot transfer learning task (``Om-Trn'' column in Table \ref{table:rec-mse}). We also considered classification as the APC task but found  reconstruction to be more suited to part-whole learning (see Supplementary Materials).

\begin{table}
\center
\begin{tabular}{|l|l|l|l|l|l|}
\hline
   & \textbf{MNIST} & \textbf{FMNIST} & \textbf{Om-Tst}& \textbf{Om-Trn} \\\hline
\textbf{RB}    & $0.0120$ & $0.0145$ & $0.0307$  & $0.0301$     \\ \hline
\textbf{APC-1}& $0.0114$ & ${\bf 0.0138}$ & $0.0324$  & $0.0323$     \\ \hline
\textbf{APC-2}      & ${\bf 0.0085}$ & ${\bf 0.0138}$ & ${\bf 0.0227}$  & ${\bf 0.0226}$    \\ \hline
\end{tabular}
%\vspace*{0.1in}
\caption{\textbf{Ablation Studies: Reconstruction Mean-Squared-Error (per pixel) for Different Models Across Datasets:} See text for details. FMNIST, Om-Tst and Om-Trn denote Fashion-MNIST, the Omniglot test and transfer datasets respectively.}
\label{table:rec-mse}
\end{table}

\vspace*{-0.05in}
\section{APPLICATION TO HIERARCHICAL PLANNING}
\label{maze}
\vspace*{-0.1in}
We now show that the same APC framework used for learning part-whole hierarchies can also be used for learning hierarchical world models for efficient planning. We introduce a new compositional, scalable ``multi-rooms'' building navigation task to illustrate this. The APC model's state-action hierarchy lends itself naturally to other RL problems as well, which we hope to demonstrate in future work. 

 \vspace*{-0.1in}
\subsection{APC Solution to Hierarchical Planning}
\begin{figure}[t]
\centering
\includegraphics[width=0.7\linewidth]{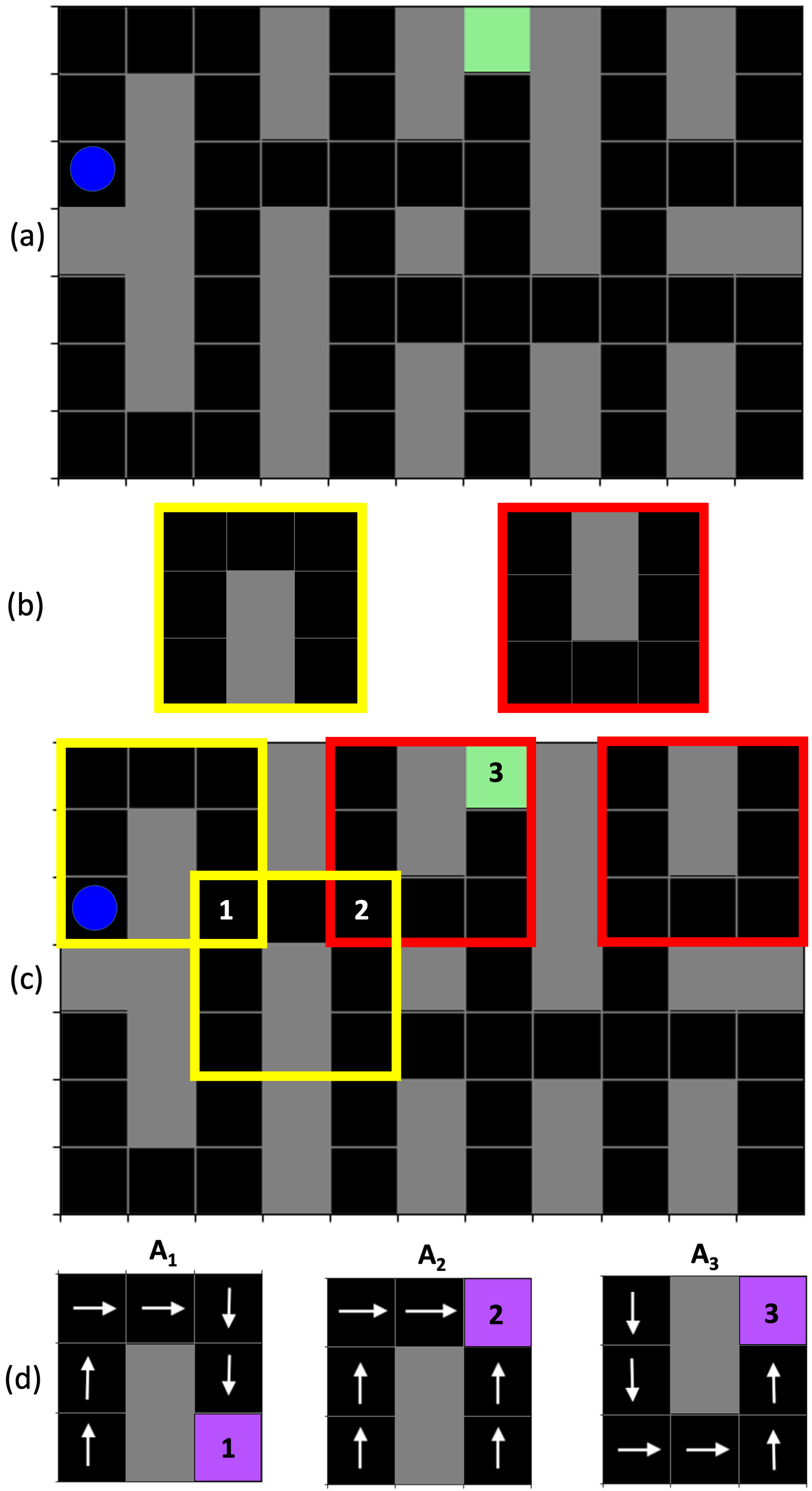}
\caption{
\textbf{The Multi-Rooms Building Navigation Problem and State-Action Hierarchy}. The problem of navigating in a large building (blue: agent location, gray: walls, green: goal) can be reduced to planning using high-level states ((b) and (c)) and actions (d). See text for details.
}
\label{fig:large-maze}
\end{figure}

\vspace*{-0.1in}
Consider the problem of navigating from any starting location to any goal location in a large ``multi-rooms'' building environment such as the one in Figure~\ref{fig:large-maze} (a) (gray: walls, blue circle: agent, green square: current goal). In the traditional non-hierarchical RL approach, the states are the discrete locations in the grid, and actions are going north (N), east (E), south (S) or west (W). A large reward (+10) is received at the goal location, with a small negative reward (-0.1) for each action to encourage shortest paths.

{\bf Problems with traditional RL}. There are three major problems with the traditional RL approach: (1) Sample inefficiency: As the environment gets larger, the number of interactions with the environment required to learn the value function explodes, (2) Risk of catastrophic consequences: Taking actual actions in the real-world to estimate the value function might have catastrophic consequences (injury or death), and (3) Inflexibility: Even after learning the value function to reach a particular goal, any change in the goal requires learning a whole new value function. 

{\bf How the APC model solves these problems.} First, note that just as an object (e.g., an MNIST digit in Section~\ref{vision}) consists of the same parts (e.g., strokes) occurring at different locations, the multi-rooms environment in Figure~\ref{fig:large-maze} is also made up of the same two components (``Room types'' R1 and R2), shown in Figure~\ref{fig:large-maze} (b), occurring at different locations (some example locations highlighted by yellow and red boxes in Figure~\ref{fig:large-maze} (c)). These components form part of the higher-level states in the APC and are defined by state embedding vectors R1 and R2, which can be trained to generate, via the hypernet $H_s$ (Figure~\ref{fig:main}), the lower-level transition functions $f_s$ for rooms R1 and R2 respectively.

Next, similar to how the APC model was able to reconstruct an image in Section~\ref{vision} using higher-level action embedding vectors to generate policies and actions (locations) to compose parts using strokes, the APC model can compute higher-level action embedding vectors ${\bf A}_i$ (option vectors) for the multi-rooms world that generate, via hypernet $H_a$ (Figure~\ref{fig:main}), lower-level policies $f_a$ which produce primitive actions (N, E, S, W) to reach a goal $i$ encoded by ${\bf A}_i$. 

{\bf Local reference frames allow policy re-use and transfer.} Figure~\ref{fig:large-maze} (d) illustrates the bottom-level policies for three such action embedding vectors ${\bf A}_1$, ${\bf A}_2$ and ${\bf A}_3$, which generate policies for reaching goal locations 1, 2, and 3 respectively. Note that the ${\bf A}_i$ are defined with respect to higher-level state R1 or R2. Defining these policies to operate within the local reference frame of the higher-level state R1 or R2 (regardless of global location in the building) confers the APC model with enormous flexibility because the {\em same policy can be re-used at multiple locations} to solve local tasks (here, reach sub-goals within R1 or R2). For example, to solve the navigation problem in Figure~\ref{fig:large-maze} (c), the APC model only needs to plan and execute 3 higher-level actions or options: ${\bf A}_1$ followed by ${\bf A}_2$ followed by ${\bf A}_3$, compared to planning a sequence of 12 lower-level actions to reach the same goal. Finally, since the ${\bf A}_i$ embedding space of options is continuous, there is an unprecedented opportunity for the APC model to exploit properties of this space (such as smoothness) to interpolate or extrapolate to create and explore new options for transfer learning; this possibility will be explored in future work.

\vspace*{-0.1in}
\subsection{Results}
\begin{figure}[t]
\centering
\includegraphics[width=0.93\linewidth]{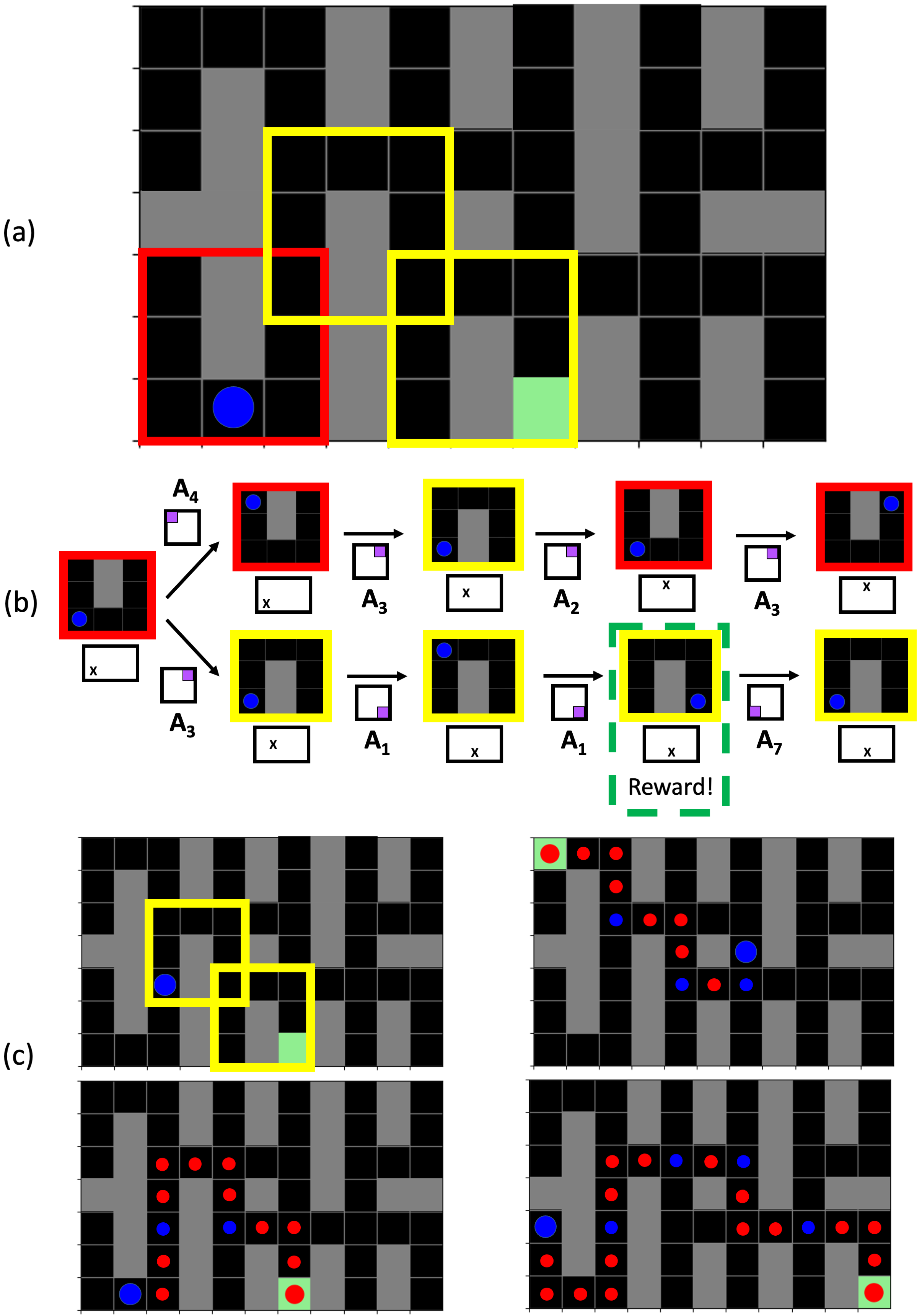}
\caption{
\textbf{Planning and Model Predictive Control}. To navigate to the green goal in (a), the agent (blue) uses its learned high-level state network to sample $N$ high-level state-action trajectories ($N=2$ in (b)), picks the sequence with highest total reward, executes this sequence's first action to reach the location in (c) (top left), and repeats to reach the goal with 3 high-level actions (bottom left in (c)). Two more planning examples are shown on the right in (c) (small red dot: intermediate location; small blue dot: intermediate goal). In (b), high-level state is depicted by predicted R1 or R2 image and its location X in global frame, action by its goal (purple) in a square local frame.  }
\label{fig:planning-ex}
\end{figure}

\begin{figure}
\centering
\begin{subfigure}{\linewidth}
 \centering
  \includegraphics[width=0.8\linewidth]
  {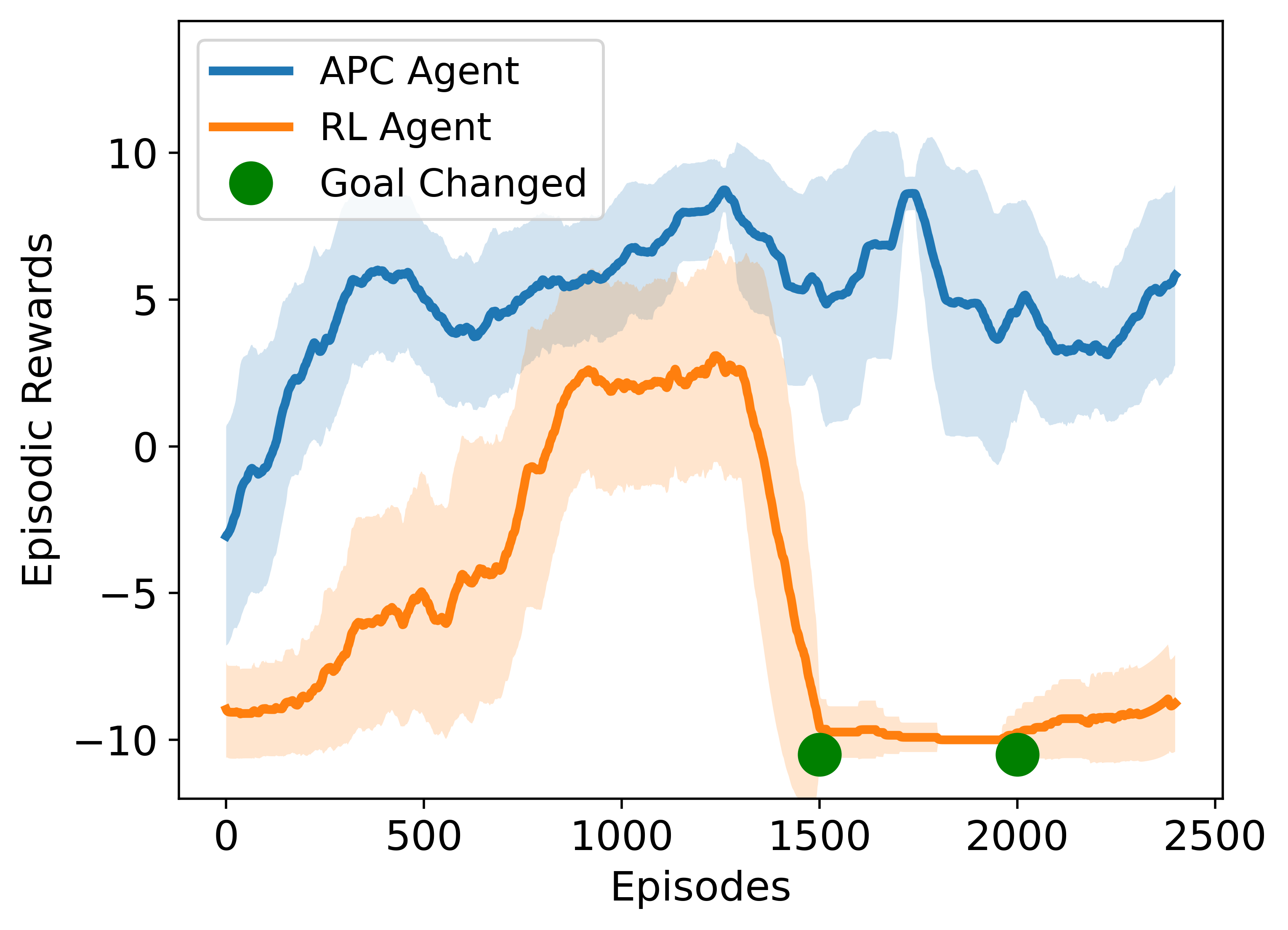}
    \caption{}
    \label{fig:multigoal}
\end{subfigure}
~
\begin{subfigure}{\linewidth}
 \centering
  \includegraphics[width=0.8\linewidth]
  {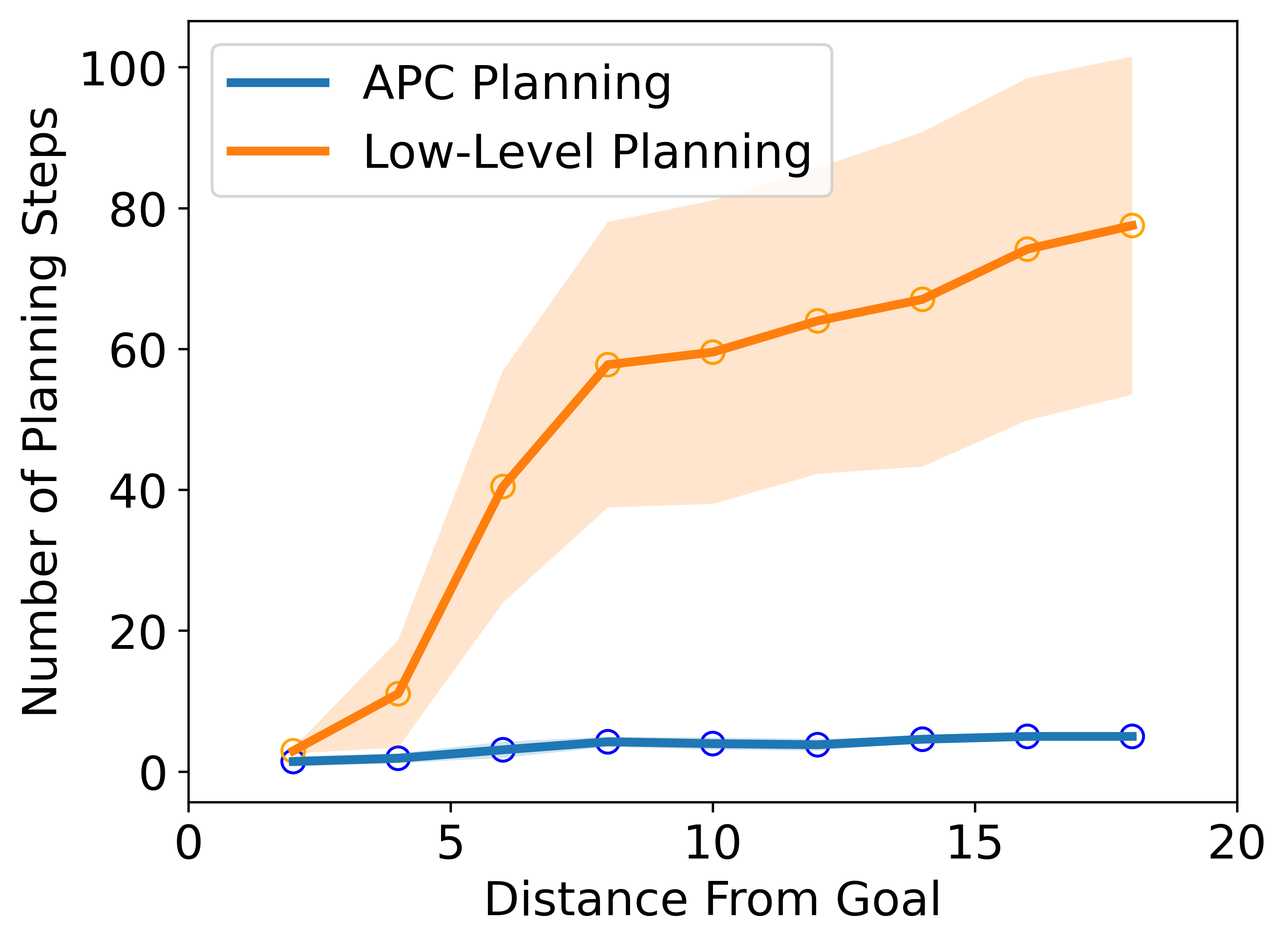}
    \caption{}
    \label{fig:planning-performance}
\end{subfigure}
\caption{
\textbf{APC Model Performance.} See text for details.
}
\label{fig:RL-plot}
\end{figure}

\vspace*{-0.1in}
We used a simple curriculum to train the two-level APC model as follows.\\
{\bf States}. For simplicity, we assume the higher-level states capture $3\times 3$ local reference frames and are defined by an embedding vector generating the transition function for ``room type'' R1 or R2 , along with the location for this local reference frame in the global frame of the building.\\
{\bf Actions}. The lower-level action network is trained to map a higher-level action embedding vector ${\bf A}_i$ to a lower-level policy that navigates the agent to a particular goal location $i$ within R1 or R2. For the current example, eight embedding vectors ${\bf A}_1,\ldots,{\bf A}_8$ were trained, using REINFORCE-based RL \citep{REINFORCE-ref}, to generate via the hypernet $H_a$ eight lower-level policies to navigate to each of the four corners of room types R1 and R2.\\
{\bf Prediction and Planning}. The higher-level state network $F_s$ was trained to predict the next higher-level state (decoded as an image of room type R1 or R2, plus its location) given the current higher-level state and higher-level action. The trained higher-level state network $F_s$ was used for planning at each step a sequence of 4 higher-level actions using random-sampling shooting model-predictive control (MPC) \citep{random-samp-shooting-ref}: random state-action trajectories of length 4 were generated using $F_s$ by starting from the current state and picking one of the four random actions $A_i$ for each next state; the action sequence with the highest total reward was selected and its first action was executed. Figure~\ref{fig:planning-ex} shows an example of this high-level planning and MPC process using the trained APC model.\\
{\bf Performance Comparison with Traditional RL and Planning  on Goal-Changing Navigation Task}. We compared the two-level APC model with both a heuristic lower-level-only planning algorithm and a REINFORCE-based RL algorithm using primitive states and actions. The task involved navigating to a randomly selected goal location in a building environment (as in Figure~\ref{fig:large-maze} (a)), with the goal location changing after some number of episodes. Figure~\ref{fig:RL-plot} (a) shows how the APC model, after an initial period spent on learning the hypernet $H_a$ to generate the lower-level options, is able to cope with goal changes and successfully navigate to each new goal by sequencing high-level actions (+10 reward for goal; -0.1 per primitive action). The RL algorithm, on the other hand, experiences a drop in performance after a goal change and does not recover even after 500 episodes. Figure~\ref{fig:RL-plot} (b) demonstrates the efficacy of APC's higher-level planning compared to lower-level planning (MPC using random sequences of 4 primitive future actions; Euclidean distance heuristic; see Supplementary Materials for details): the average number of planning steps to reach the goal increases dramatically for larger distances from the goal  for the lower-level planner compared to higher-level planning by the APC model.

\vspace*{-0.05in}
\section{CONCLUSION}
\vspace*{-0.1in}
\label{sec:conclusion}
We have presented, to our knowledge, the first demonstration of a unified solution to the part-whole learning problem \citep{DBLP:journals/corr/abs-2102-12627}, the nested reference frames problem \citep{hawkins2021thousand}, and the integrated state-action hierarchy learning problem in reinforcement learning \citep{HRL-review}. Our APC model is inspired by the growing evidence for predictive sensory-motor processing in the cortex \citep{Schneider2018,KELLER2018424}. The APC model for equivariant vision employs ``eye movements" for visual sampling, and performs end-to-end learning and parsing of part-whole hierarchies from images. 
%The incremental parsing of a scene by our model using state-action recurrent networks also provides an explanation for why our visual perception is stable despite the staccato nature of our eye movements. 
We also showed how the same APC framework provides a flexible approach to hierarchical planning and action selection. Given the growing interest in predictive coding as a model for the cortex, our results showing two very different  applications of the same APC model lend further support to the hypothesis of a possible common computational principle operating across cortex  \citep{Mountcastle}. Current limitations of the model include the fixed number of time steps used at each level and using only a two-level  hierarchy. Future work will explore deeper versions of the model, comparisons to neuroscience data, and more sophisticated planning/RL methods to enable the model to scale to more complex tasks. 

%The framework we have proposed is quite general and flexible. For example, actions in APCNs could include not just position but arbitrary transformations of parts, allowing the network to learn hierarchical equivariant representations  \citep{DBLP:journals/corr/abs-2102-12627,DBLP:conf/iclr/HintonSF18}. More broadly, the APCN framework offers a new approach to modeling and solving hierarchical reinforcement learning and planning problems in continuous state and action spaces. The APCN model also generalizes the dynamic predictive coding model \citep{Jiang2021DynamicPC} to sensory-motor learning at multiple timescales and offers a new role for cortical feedback connections in modulating the dynamics of lower-level sensory and motor networks.  

\newpage

\setcounter{section}{0}
\setcounter{figure}{0}
\setcounter{footnote}{0}
\setcounter{table}{0}

\definecolor{codegreen}{rgb}{0,0.6,0}
\definecolor{codegray}{rgb}{0.5,0.5,0.5}
\definecolor{codepurple}{rgb}{0.58,0,0.82}
\definecolor{backcolour}{rgb}{0.95,0.95,0.92}

\lstdefinestyle{mystyle}{
    backgroundcolor=\color{backcolour},   
    commentstyle=\color{codegreen},
    keywordstyle=\color{magenta},
    numberstyle=\tiny\color{codegray},
    stringstyle=\color{codepurple},
    basicstyle=\ttfamily\footnotesize,
    breakatwhitespace=false,         
    breaklines=true,                 
    captionpos=b,                    
    keepspaces=true,                 
    numbers=left,                    
    numbersep=5pt,                  
    showspaces=false,                
    showstringspaces=false,
    showtabs=false,                  
    tabsize=2
}

\lstset{style=mystyle}
\setlength{\footskip}{3.30003pt}
\setlength{\topskip}{-5pt}

\onecolumn
\aistatstitle{Supplementary Materials\\
\vspace*{0.1in}Active Predictive Coding: A Unified Neural Framework for Learning Hierarchical World Models for Perception and Planning
}

%\vspace*{-0.5in}
\section{Active Predictive Coding for Equivariant Vision}
\subsection{Experimental Details}
\label{app-exp-details}

\subsubsection{Parameter Settings}
For the active predictive coding (APC) model for vision introduced in Section 5, for all datasets (MNIST, FMNIST, Omniglot and affNIST\footnote{affNIST is only used for the comparison in Supplementary Materials Section~\ref{app:hyp-vs-emb}.}), the top-level action and state RNN activity vectors were of size $256$, while the lower level ones were of size $32$ for MNIST, Fashion-MNIST and $64$ for Omniglot. Both hypernetworks $H_s$ and $H_a$ consisted of four layers with sizes $256,256,64$ and $|\theta_s|$ or $|\theta_a|$. The last two layers had linear activation functions. Since the number of units of the last layer was equal to the number of parameters of $f_a$ or $f_s$, the middle layer functions as a bottleneck layer. $F_a$ and $F_s$ were implemented as RNNs with structure similar to the network used in RAM \citep{NIPS2014_09c6c378}. The option vectors ${\bf z}$ were of size $32$. RELU activations were employed throughout the model apart from the last level of the reconstruction network (to avoid ``dead'' pixels). The glimpse scales were set such that $I^{(1)}$ was $14\times14$ and $g_{t,\tau}$ $7\times7$ pixels. The sizes of MNIST and Fashion-MNIST images are $28\times 28$ pixels, whereas we downsampled Omniglot characters to $32\times 32$ pixels.

For Omniglot, we reserved $85\%$ (rounded down) of the character classes in each alphabet as part of our training set. The rest of the character classes were used as the transfer set. Within the training classes, we reserved $3$ examples from each character class for the validation set and another $3$ for the test set.

\subsubsection{Initialization}
We utilize random glimpses to initialize the top-level state and action vectors. A random glimpse is generated at location $l_\text{init} \sim \mathcal{U}[-0.5,0.5]$. This initialization glimpse $g_\text{init}$, together with a small trainable initialization network, initializes the state vector ${\bf R}_{0}$. The action vector is initialized using $F_a$ and ${\bf R}_{0}$ by setting the previous action vector and feedback $\rho_A$ to all-zeroes vectors.

\subsubsection{Training Settings}

For training, we used the Adam optimizer \citep{DBLP:journals/corr/KingmaB14}. For all datasets, we trained the model for a number of epochs with a learning rate of $0.0001$ (first phase). Then we reduced the learning rate to $0.00001$ (second phase). The number of epochs for each phase and dataset are shown in Table \ref{tab:epoch-settings}.

\begin{table}[h!]
\center
\begin{tabular}{|l|l|l|}
\hline
   & \textbf{First phase} & \textbf{Second phase} \\\hline
\textbf{MNIST}& $1000$ & $200$ \\ \hline
\textbf{FMNIST}& $1000$ & $200$ \\ \hline
\textbf{Omniglot}& $10000$ & $2000$ \\ \hline
\textbf{affNIST}& $2000$ & $500$ \\ \hline
\end{tabular}

\caption{\textbf{Training schedules for each dataset.}}
\label{tab:epoch-settings}
\end{table}

\subsubsection{Location Penalty}
\label{app:penalty}
\begin{figure}[h]
\centering
\includegraphics[scale=0.4]{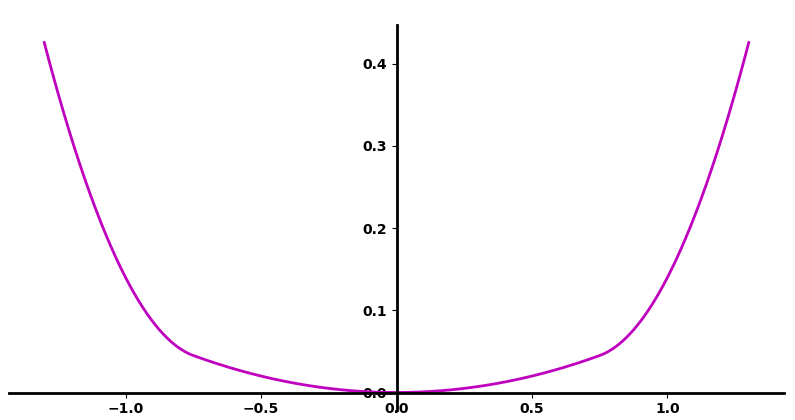}

\caption[APC location penalty function for $c=0.75$]{\textbf{APC location penalty function for $c=0.75$.}}
\label{fig:app-penalty}
\end{figure}

Ideally the APC model for vision in Section 5 would avoid generating locations exceeding the boundaries of the image. Several implementations of RAM \citep{NIPS2014_09c6c378} use clipping or the hyperbolic tangent activation function. In practice, we found that constraining the locations via an appropriate penalty was more effective. We calculate a threshold $c$ so that if a glimpse is centered $c$ units away from the image boundary ($l\in[-1.0+c,1.0-c]$), then the glimpse resides entirely within that boundary. We then apply a thresholded version of $\ell_2$ normalization:
$$
L_\text{reg}(l) = (\text{LRelu}(l-c))^2 + (\text{LRelu}(-l-c))^2 - 2(\alpha c)^2
$$
where LRelu is the leaky rectified linear unit with $\alpha=0.2$. The structure of this penalty can be seen in Figure \ref{fig:app-penalty}.

\subsection{Using Classification instead of Reconstruction as the APC Task}
We initially considered classification as a task for the APC model. A previous model  \citep{NIPS2014_09c6c378} showed that using a single glimpse randomly sampled from the entire input image achieves $57.15\%$ classification accuracy on the MNIST dataset. We found that our RB baseline with three random glimpses already achieved $93.1\%$ classification accuracy. This suggested  to us that rather than classification (where random glimpses appear to give good accuracies at least for the MNIST dataset), the task of reconstructing an object, such as an MNIST digit, might be a more appropriate task for learning and enumerating the parts of an object. We therefore chose image reconstruction as the task for testing the APC model. 

\subsection{Hypernetworks versus Traditional Embedding-Based Alternate Model}
\label{app:hyp-vs-emb}

\begin{table}[h!]
\center
\begin{tabular}{|l|l|}
\hline
   & \textbf{affNIST} \\\hline
\textbf{APC-2-Embedding}& $0.0110$ \\ \hline
\textbf{APC-2-Hypernet}      & ${\bf 0.0105}$ \\ \hline
\end{tabular}

\caption{\textbf{Hypernetworks vs.\ Embedding-Based Alternative for Generating Sub-Programs for affNIST.}}
\label{table:rec-ablation}
\end{table}
An alternative to using hypernetworks to generate the state and action sub-programs is to use an embedding-based approach \citep{NEURIPS2020_75c58d36}: the higher-level embedding vector is fed as input to a feedforward network and its output is concatenated with the input to the bottom level RNN.
We compared the two approaches on the reconstruction task for affNIST, a dataset that consists of MNIST digits embedded in a $40\times 40$-pixel frame and transformed via random affine transformations. The results, shown in Table \ref{table:rec-ablation}, indicate that hypernetworks provide a more efficient abstraction for sub-program generation. See \citep{NEURIPS2020_75c58d36} for a theoretical discussion of the advantages of hypernetworks over the embedding approach.

\section{Active Predictive Coding for Hierarchical Planning} \label{section:apc_planning}

\subsection{APC Model Components}
To test hierarchical planning, we introduce a new task involving navigation in a compositional environment (as shown in Figure 8 of the paper). The complete APC model for this task can be decoupled into independent trainable components corresponding to each level of a hierarchy. For training data, we collect the state-action tuples $({\bf r}_{\tau}, {\bf a}_{\tau}, {\bf r}_{\tau+1})$ and $({\bf R}_t, {\bf A}_t, {\bf R}_{t+1})$ for each episode (see Figure 2 from the paper for details) and use it to train lower-level networks ($f_s$, $f_a$) and higher-level networks ($F_s$, $F_a$) respectively.

\subsubsection{Lower-Level Networks}
The lower networks ($f_s$, $f_a$) are generated by hypernetworks ${H_s}$ and ${H_a}$ in Figure 2 in the paper. In the comparison with the APC hierarchical planner (see below), $f_s$ was chosen to be the ground truth environment transition model to give the lower-level planner the best possible advantage when comparing to the APC planner. The parameters $\theta_a$ of action policy network $f_a$ was generated from the hypernetwork $H_a$ that takes the abstract action vector ${\bf A}_i$ as input. $H_a$ is initialized with principled weight initialization strategies that have proven benefits for hypernetworks \citep{Chang2020Principled}. Our hypernetwork is a simple 4-layer dense neural net with ReLU activations. ${H_a}$ outputs parameters for a 3-layer policy network $f_a$. This policy network allows sampling of actions in an on-policy fashion, and generates the tuples $({\bf r}_t, {\bf a}_t, G_{t}, {\bf r}_{t+1})$ where $G_t$ is the discounted reward. The entire model is trained using the episodic REINFORCE algorithm \citep{NIPS1999_464d828b,REINFORCE-ref}. Baseline methods are used to reduce variance in the gradient computation. We also apply an $L_2$ regularization constraint on the output action logits, which has been shown to be highly beneficial in improving the stability of training. Formally,

\[ \theta_{a, i} = {H_a}({\bf A}_i) ;\quad   {\bf a}_t \sim f_a({\bf r}_t ; \theta_{a, i})\]
\[ L_{REINFORCE} = J(\theta_{H_a}, \theta_{a, i}) = \sum_{t=1}^{T_1} \left[\underbrace{-\log\ \pi({\bf a}_{t} | {\bf r}_{t}; f_a) (G_{t} - b({\bf r}_{t}))}_{Policy\ Gradient\ Loss} + \underbrace{|| f_a({\bf r}_t;\theta_{a, i} )||_2^2}_{L_2 \ Loss} \right]\]

Here $T_1$ is the length of the episode and $b({\bf r}_t)$ is the baseline value for the state ${\bf r}_t$. We note that the APC architecture is independent of the RL algorithm used to train the model. More efficient Actor-Critic and DDPG methods will be explored in our future work.

\subsubsection{Higher-Level Networks}
To illustrate the performance of the APC model on navigation tasks, we choose eight higher-level actions (${\bf A}_1, {\bf A}_2, ..., {\bf A}_8$), one for each subgoal within the two room types $R_1$ and $R_2$. Each vector ${\bf A}_i$ is a one-hot vector representing this subgoal. To perform hierarchical planning, we learn the higher-level state transition network $F_s$ which predicts the next higher-level state ${\bf R}_{t+1}$ given the current higher-level state ${\bf R}_{t}$ and action ${\bf A}_t$. The network $F_s$ receives as input the concatenated representation containing room location and an image of the room, and predicts the same representation for the next higher-level time step. We implement the recurrent network $F_s$ as a simple 4 layer neural network with one recurrent LSTM layer, with ReLU and eLU activations. The  reconstruction error (mean square error) between predicted and actual inputs was used as the loss function to train $F_s$ via gradient descent and backpropagation.

The higher action policy $F_a$ which outputs ${\bf A}_{t}$ given ${\bf R}_{t}$ and ${\bf A}_{t-1}$ can be trained with the planned action sequences that were successful in navigating to the goal. This will be explored in future work.

\subsection{Experimental Details}

\begin{table}[t]
\centering

\catcode`,=\active
\def,{\char`,\allowbreak}

\begin{tabular}{p{8.4cm}<{\raggedright} p{6cm}<{\raggedright} }
    \toprule
    \textbf{Hyperparameters}      & \textbf{Selected Value} \\ 
    \midrule
    
    Agent Learning Rate ($\alpha_a$)        & $1e-3$ \\
    Baseline Learning Rate ($\alpha_b$)     & $1e-3$ \\
    Higher Level State Model Learning Rate ($\alpha_{hl}$)  & $5e-4$  \\ 
    Regularization Coefficient ($\lambda_{L2}$)  & $5e-3$ \\
    Discount Factor ($\gamma$)              & $0.99$ \\
    Hypernetwork Layers ($H_a$)             & $[16, 128, 128, *]$  \\
    Policy Network Layers ($f_a$)           & $[64, 64, 4]$ \\
    Higher Level State Model Layers ($F_s$)              & $[128, 256, 256]$ \\
    Activation functions used               & ReLU and its variants \\
    Optimizer                               & Adam   \\ 
  \bottomrule
   
\end{tabular} 

\caption{{\bf Model Hyperparameters used to obtain the APC Hierarchical Planning Results in the Paper.} These were the best set of configurations among various combinations of hyperparameters that were explored. The star(*) indicates that the final layer of hypernetwork produces weights of each layer of the policy network independently.}
\label{tb:hyperparam_table}
\end{table}

Table \ref{tb:hyperparam_table} lists the important hyperparameters used to train our APC model for planning. The code snippet to train the hypernetwork agent is provided in Listing \ref{lst:hyper_code}. The goal of our planning experiments was to show that (1) APC model can be extended to navigation and RL domains, (2) APC is robust to changes in goals and (3) the hierarchy can be exploited to plan at exponentially faster rates. Here, we discuss the experimental setup we used to help achieve these goals.

\subsubsection{Comparison with Traditional RL Agent}

To illustrate the APC model's zero-shot planning ability in a goal-changing navigation task, we compared the APC model to a traditional RL agent with no hierarchy. The RL agent is a relatively large recurrent LSTM neural net with layers $[128, 256, 128, 32, 4]$. This neural net acts as a policy and takes current primitive states (locations) $s_t$ as input. Primitive actions are sampled from the action probabilities output by this network. This agent is trained via REINFORCE (with baseline), similar to the lower level policy networks in the APC model, in an online and episodic fashion. 

During training, the goal was changed when the  RL agent had learned a reasonable policy (see Figure 10 (a) in the paper). Unsurprisingly, the episodic rewards drop rapidly with the change in goals. This is not the case for our APC agent, which learns an abstract world model and plans in this abstract space. With simple planning strategies like Model Predictive Control, the APC agent can solve the task with dynamically changing goals, as observed in Figure 10 (a) in the paper.

\subsubsection{Comparison with Lower-Level Planner}

The APC agent also leverages hierarchy to plan in fewer numbers of steps. To demonstrate this, we compare against a lower-level planner. To provide the lower-level planner with the best advantage possible in competing with the APC model, we assume that the lower-level planner's state transition model is the underlying ground truth transition function $f_s$  capturing the transition from current to next state given an action in the environment. The lower-level planner used a Euclidean distance heuristic to get a rough estimate of a given state's distance to goal at each planning step.

Model Predictive Control (MPC) was used to plan in both the APC and lower-level planner. We used 4-step look-ahead with random action sequences in both cases. For the APC model, we used the learned transition model $f_s$ for the lower level and $F_s$ for the higher level. $F_s$  was unrolled with the random action sequence (${\bf A}_t, {\bf A}_{t+1}, {\bf A}_{t+2}, {\bf A}_{t+3}$). $N$ such random trajectories were unrolled and the sequence reaching the goal was chosen. If no such sequence was found, a random sequence was chosen and the first action was performed. This process was repeated until the goal was reached and the number of such planning steps was recorded. Note that the APC planner in this experiment had no access to heuristics that would inform the location of goal and relied only on feedback regarding whether an action sequence reached the goal.

The lower-level planner followed a similar MPC strategy, but with the added advantage of having access to the true underlying transition model and heuristics to bias action selection towards the actions that improve the heuristic measure. Even with these advantages, we found that the APC model plans dramatically faster than the lower-level planner, as seen in Figure 10 (b). This suggests that the state-action hierarchy in the APC model has the potential to greatly improve sample efficiency and exploration.

\subsection{Code Snippet for Training a 2-Level APC Agent} 
%% This declares a command \Comment
%% The argument will be surrounded by /* ... */
% \RestyleAlgo{ruled}
% \SetKwComment{Comment}{/* }{ */}
% \setlength{\footskip}{3.30003pt}
% \setlength{\algomargin}{1em}
% \begin{algorithm}

% \caption{Training the Lower APC Agent via Hypernetworks}\label{alg:two}

% \KwIn{$A_i$, \texttt{environment}}
% \KwOut{\texttt{lower\_agent}}

% $y \gets 1$\;
% $X \gets x$\;
% $N \gets n$\;
% \While{$N \neq 0$}{
%   \eIf{$N$ is even}{
%     $X \gets X \times X$\;
%     $N \gets \frac{N}{2}$ %\Comment*[r]{This is a comment}
%   }{\If{$N$ is odd}{
%       $y \gets y \times X$\;
%       $N \gets N - 1$\;
%     }
%   }
% }
% \end{algorithm}
\begin{lstlisting}[language=Python, label={lst:hyper_code}, caption={\bf Code Snippet for Training the Hypernetwork-Modulated APC Agent (}$H_a$ {\bf and} $f_a${\bf ).}] 
import tensorflow as tf
import tensorflow_probability as tfp

def forward_pass(environment, abstract_action):
    
    # The hypernetwork is used to instantiate parameters of a policy, specific to the A informed by higher layers.
    policy_parameters = generate_policy_parameters(abstract_action)
    
    # A dynamic policy is generated and used to navigate in the environment
    policy = lambda x: construct_policy(x, policy_parameters)
    
    collect_rewards, collect_log_probs, collect_l2 = [], [], []
    
    for i in range(MAX_TIMESTEPS):
        
        # Forward pass with the policy
        state = environment.get_current_state()
        action_logits = policy(state)
        
        # Sample action and compute their probabilities
        distribution = tfp.distributions.Categorical(logits=action_logits)
        action = distribution.sample()
        log_probs = distribution.log_prob(action)
        l2 = tf.reduce_sum(tf.square(action_logits))
        
        # Take a step in the environment
        next_state, reward, done, _ = environment.step(action)
        
        collect_rewards.append(reward)
        collect_log_probs.append(log_probs)
        collect_l2.append(l2)
        
        if done:
            break
    
    # Compute loss using policy gradients
    reinforce_loss = REINFORCE(collect_rewards, collect_log_probs, collect_l2)
    
    return reinforce_loss

\end{lstlisting}

%\subsection{Hyperparameters Table}

\subsection{Additional Results: Zero-Shot Transfer to New Environments}

\begin{figure}[h]
\centering
\includegraphics[width=0.9\textwidth]
  {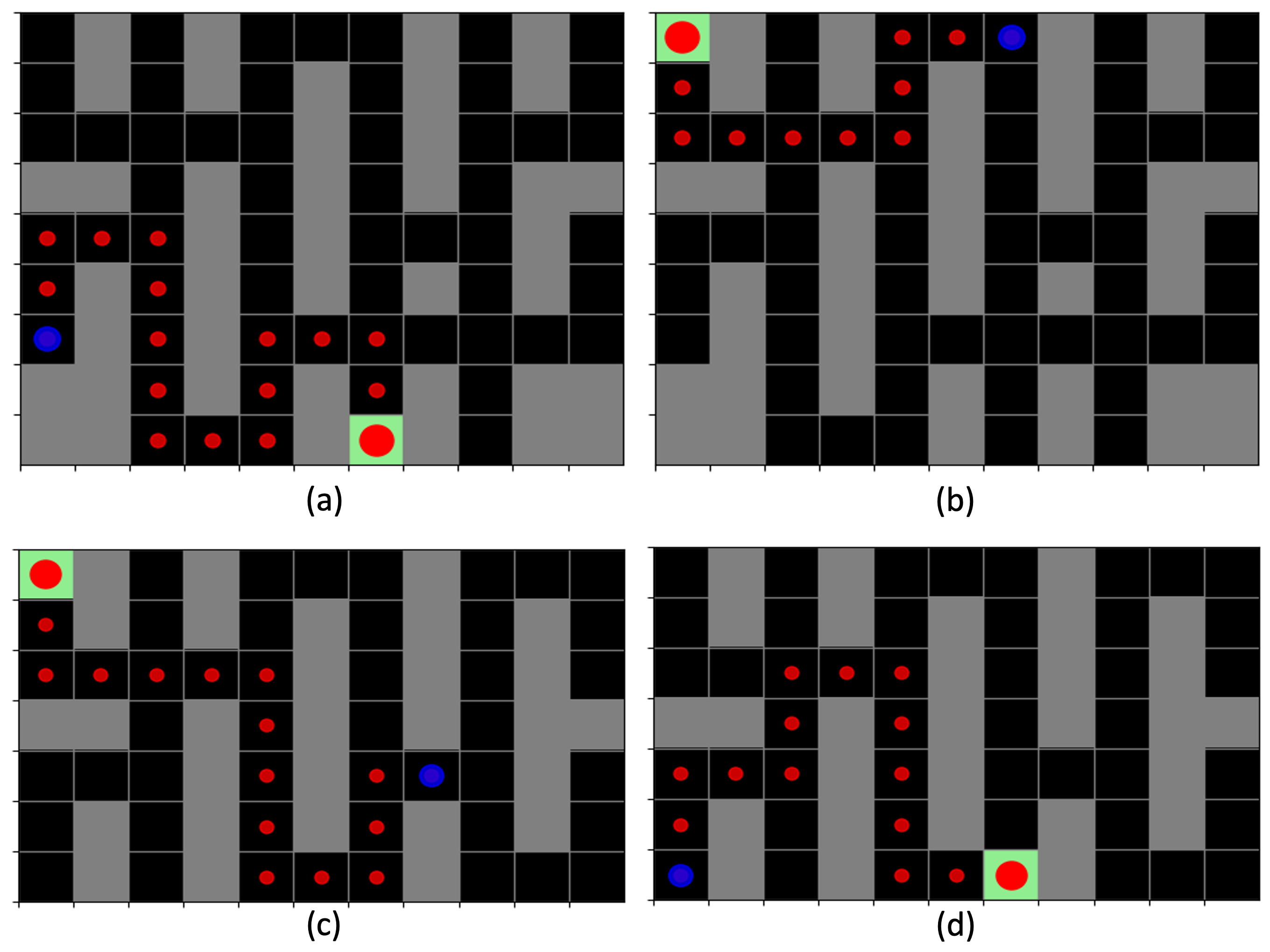}
% \begin{subfigure}{0.5\textwidth}
%  \centering
%   \includegraphics[height=0.2\textwidth]
%   {figs/planned_path_1.png}
%     \caption{}
%     \label{fig:planning_new_1}
% \end{subfigure}
% ~
% \begin{subfigure}[t]{\linewidth}
%  \centering
%   \includegraphics[height=0.2\textwidth]
%   {figs/planned_path_2.png}
%     \caption{}
%     \label{fig:planning_new_2}
% \end{subfigure}
% ~
% \begin{subfigure}{\linewidth}
%  \centering
%   \includegraphics[height=0.25\textwidth]
%   {figs/planned_path_3.png}
%     \caption{}
%     \label{fig:planning_new_3}
% \end{subfigure}
% ~
% \begin{subfigure}{\linewidth}
%  \centering
%   \includegraphics[height=0.25\textwidth]
%   {figs/planned_path_4.png}
%     \caption{}
%     \label{fig:planning_new_4}
% \end{subfigure}
\caption{\textbf{Zero-Shot Transfer to New Environments.} (a)-(d) show hierarchical planning by the APC model using the same set of higher-level actions as in the paper to solve new navigation problems in new  environments. Each row is a different new environment. Blue dot denotes a start state. Navigation (red dots) to two randomly chosen goals (green) are shown for each new environment.
}
\label{fig:planning_big}
\end{figure}

We also tested the APC model on navigation problems in a variety of other large compositional building environments. The same eight higher-level ``macro-actions'' (options) learned for the environment in Figure 8 (a) in the paper were used by the APC model to successfully compose macro-action sequences to navigate to different goals in new environments, an example of zero-shot transfer. Example results showing hierarchical planning by the APC model in these new large environments are shown in Figure~\ref{fig:planning_big}.  

% \begin{figure}[ht]
% \centering
% \includegraphics[scale=0.6]{figs/planned_path.png}

% \caption[]{\textbf{Zero-Shot Transfer to a New Environment.} The image shows hierarchical planning by the APC model using the same set of higher-level actions as in the paper to solve a new navigation problem in a new complex environment.}

% %\caption[]{\textbf{Zero-Shot Transfer to New Environments.} The images show hierarchical planning by the APC model using the same set of higher-level actions as in the paper to solve new navigation problems in new complex environments. Each row is a different environment. Navigation to two randomly chosen goals are shown for each new environment. }
% \label{fig:planning_big}
% \end{figure}

\newpage
%\small
\bibliography{main.bib}

\end{document}